\documentclass{article}

\usepackage{microtype}
\usepackage{graphicx}
\usepackage{subfigure}
\usepackage{booktabs} %

\usepackage{rotating}
\usepackage{afterpage}

\usepackage{hyperref}

\usepackage[accepted]{icml2024}

\usepackage{amsmath}
\usepackage{amssymb}
\usepackage{mathtools}
\usepackage{amsthm}

\usepackage{amsfonts}
\usepackage{wrapfig}
\usepackage{xpatch}
\usepackage{xspace}

\usepackage{adjustbox}

\usepackage[capitalize,noabbrev]{cleveref}

\newcommand{\ourmethod}{\textsc{la\nobreakdash-nam}\xspace}

\def\1{\bm{1}}

\newcommand{\transpose}{^\mathrm{\textsf{\tiny T}}}
\newcommand{\diag}{\operatorname{diag}}

\newcommand{\inv}{^{-1}}

\newcommand{\R}{\mathbb{R}}

\def\vf{{\mathbf{f}}}

\def\vtheta{{\boldsymbol{\theta}\xspace}}

\def\mI{{\mathbf{I}}}
\def\mJ{{\mathbf{J}}}

\def\mP{{\mathbf{P}}}

\def\mS{{\mathbf{S}}}

\def\mSigma{{\boldsymbol{\Sigma}\xspace}}

\DeclareMathAlphabet{\mathsfit}{\encodingdefault}{\sfdefault}{m}{sl}
\SetMathAlphabet{\mathsfit}{bold}{\encodingdefault}{\sfdefault}{bx}{n}

\newcommand{\half}{\mbox{$\frac{1}{2}$}}

\newcommand{\trans}{^\top}
\newcommand{\E}{\mathop{\mathbb{E}}}

\newcommand{\eqdef}{\overset{\mathrm{def}}{=}}

\newcommand{\Esp}{\mathop{\mathbb{E}}} %
\newcommand{\Var}{\mathop{\mathbb{V}}} %
\newcommand{\Ent}{\mathrm{H}} %
\newcommand{\MutInf}{\mathrm{I}} %
\DeclareMathOperator{\trace}{Tr} %

\newcommand{\weights}{\boldsymbol{\theta}}
\newcommand{\inputs}{\mathbf{x}}
\newcommand{\targets}{y}
\newcommand{\allweights}{\boldsymbol{\theta}}

\newcommand{\precision}{\lambda}
\newcommand{\precisions}{\boldsymbol{\precision}}

\usepackage[textsize=tiny,textwidth=1.8cm,disable]{todonotes}
\makeatletter
\newcommand*\iftodonotes{\if@todonotes@disabled\expandafter\@secondoftwo\else\expandafter\@firstoftwo\fi}  %
\makeatother

\theoremstyle{plain}

\theoremstyle{definition}

\theoremstyle{remark}

\usepackage[textsize=tiny]{todonotes}

\icmltitlerunning{Improving Neural Additive Models with Bayesian Principles}

\begin{document}

\twocolumn[
\icmltitle{Improving Neural Additive Models with Bayesian Principles}

\icmlsetsymbol{equal}{*}

\begin{icmlauthorlist}
\icmlauthor{Kouroche Bouchiat}{eth}
\icmlauthor{Alexander Immer}{eth,mpi}
\icmlauthor{Hugo Y\`eche}{eth}
\icmlauthor{Gunnar R\"atsch}{eth}
\icmlauthor{Vincent Fortuin}{hel,tum,mcml}
\end{icmlauthorlist}

\icmlaffiliation{eth}{ETH Z\"urich, Z\"urich, Switzerland}
\icmlaffiliation{mpi}{Max Planck Institute for Intelligent Systems, T\"ubingen, Germany}
\icmlaffiliation{hel}{Helmholtz AI, Munich, Germany}
\icmlaffiliation{tum}{TU Munich, Munich, Germany}
\icmlaffiliation{mcml}{Munich Center for Machine Learning, Munich, Germany}

\icmlcorrespondingauthor{Kouroche Bouchiat}{kouroche.bouchiat@gmail.com}
\icmlcorrespondingauthor{Alexander Immer}{alexander.immer@inf.ethz.ch}

\icmlkeywords{Neural Additive Models, Bayesian Inference, Laplace Approximation, Uncertainty, Feature Selection, Feature Interaction, Automatic Relevance Determination, Machine Learning}

\vskip 0.3in
]

\printAffiliationsAndNotice{}  %

\begin{abstract}
Neural additive models (NAMs) enhance the transparency of deep neural networks by handling input features in separate additive sub-networks.
However, they lack inherent mechanisms that provide calibrated uncertainties and enable selection of relevant features and interactions.
Approaching NAMs from a Bayesian perspective, we augment them in three primary ways, namely by
    a) providing credible intervals for the individual additive sub-networks;
    b) estimating the marginal likelihood to perform an implicit selection of features via an empirical Bayes procedure; 
    and c) facilitating the ranking of feature pairs as candidates for second-order interaction in fine-tuned models.
In particular, we develop Laplace-approximated NAMs (LA-NAMs), which show improved empirical performance on tabular datasets and challenging real-world medical tasks.  
\end{abstract}

\section{Introduction}
\label{sec:introduction}

\begin{figure*}
	\includegraphics[width=\textwidth]{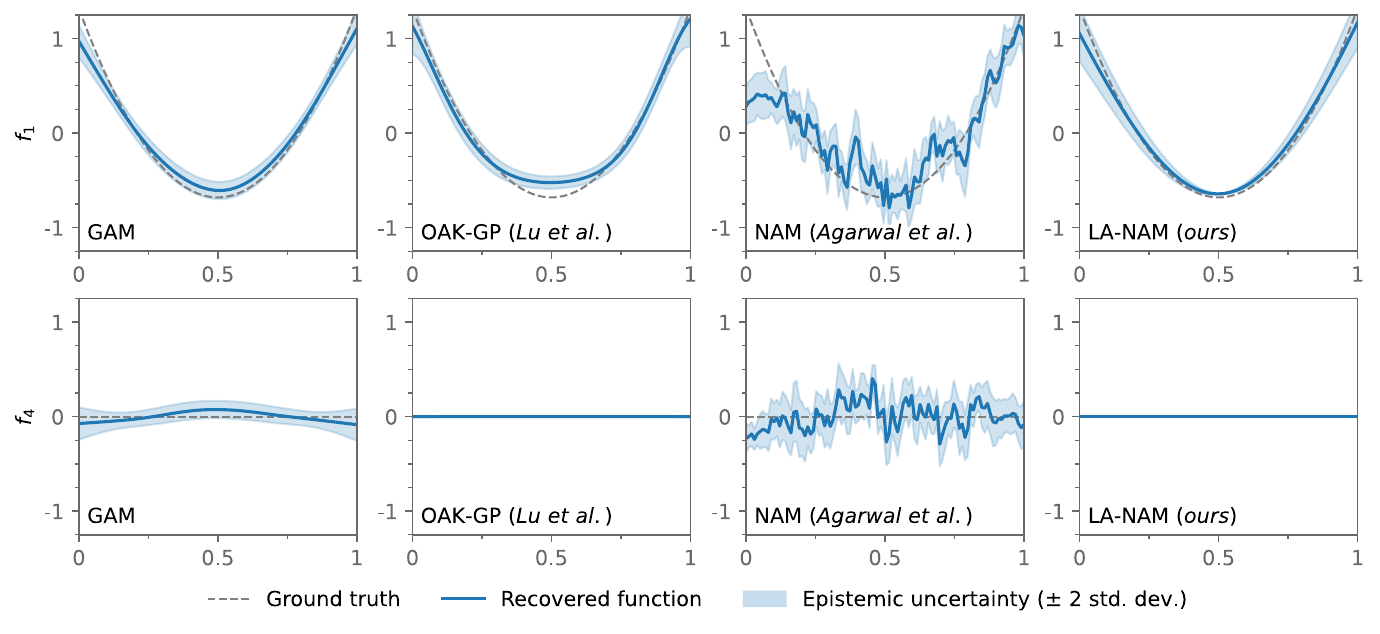}
	\caption[]{Regression on a synthetic dataset with known additive structure (see \cref{sec:exptoyexample} and \cref{apd:toyexample} for details). The \ourmethod fits the data well, provides useful uncertainty estimates, and, along with OAK-GP, correctly ignores the uninformative feature ($f_4$, bottom.)}
    \label{fig:toyexample}
\end{figure*}

Over the past decade, deep neural networks (DNNs) have found successful applications in numerous fields, ranging from computer vision and speech recognition to natural language processing and recommendation systems.
This success is often attributed to the growing availability of data in all areas of science and industry.
However, the inherent complexity and lack of transparency of DNNs have impeded their use in domains where understanding the reasoning behind their decision-making process is important \citep{pumplun2021adoption,veale2018fairness}.

Model-agnostic methods, such as partial dependence \citep{friedman2001pdp}, SHAP \citep{lundberg2017shap}, and LIME \citep{ribeiro2016lime}, provide a standardized approach to explaining predictions in machine learning, but the explanations they generate for DNNs are not faithful representations of their full complexity \citep{rudin2019stop}.
Instead, one can enhance transparency in DNNs by acting directly on the architecture and training procedure.
For instance, in generalized additive models~\citep{hastie1999gams}, the response variable $y$ is associated with inputs $x_1$, ..., $x_p$ using an additive structure of the form
\begin{equation}
  g(\E[y\,|\,x_1,\,\dots,\,x_p]) = \beta_0 + f_1(x_1) + \dots + f_p(x_p).
\end{equation}
The neural additive models (NAMs) introduced by \citet{agarwal2021nams} hinge on this concept.
In this model architecture, each input dimension is processed by a distinct sub-network, elucidating the connections between the input features and the model's predictions.
However, restricting neural network models in this way can lead to overconfidence, overreliance on uninformative features and obliviousness to  underlying feature interactions.

\paragraph{Main contributions.} In this work, we (a) propose a Bayesian variant of the NAM by deriving a tailored \emph{linearized Laplace inference} \citep{mackay1994bayesian} on the subnetworks.
We show that this improves intelligibility by \textbf{encouraging smoothness and enhancing uncertainty estimation}.
Moreover, (b) this construction enables the estimation of the marginal likelihood, which serves as a Bayesian model selection criterion.
This enables the model \textbf{to mitigate the impact of uninformative features}, further contributing to robustness and trustworthiness.
Also, (c) in situations where strong feature interaction exists in the ground-truth, causing poor performance of first-order additive models, we leverage our Laplace approximation to \textbf{automatically select informative feature interactions}, which are then added to the model to improve overall performance.
Finally, (d) we show on tabular regression and classifications benchmarks, as well as on challenging real-world medical tasks, that our proposed \ourmethod performs competitively with the baselines while offering more interpretable global and local explanations, all within a same framework.

\section{Related Work}
\label{sec:related_work}

\paragraph{Neural additive models.}

Several constructions exist for the generalized additive models (GAMs) of \citet{hastie1999gams}, including smoothing splines \citep{wahba1983smoothing} and gradient-boosted decision trees \citep{friedman2001pdp, lou2012boosting, lou2013gatwom}. Neural networks are also particularly compelling candidates for the construction of GAMs since they can be made to approximate continuous functions up to arbitrary precision given sufficient complexity \citep{cybenko1989theorem, maiorov1999theorem, lu2017theorem}.
The neural additive model (NAM) proposed by \citet{agarwal2021nams} is constructed using ensembles of feed-forward networks. In order to fit jagged functions and promote diversity for ensembling, the authors propose the exponential unit (ExU) activation, in which weights are learned in logarithmic space.
The GAMI-Net proposed around the same time by \citet{yang2021gaminet} is closely related but single networks are used in place of an ensemble, and the model also supports learning of feature interaction terms.
Recently proposed extensions to NAMs include feature selection through sparse regularization of the networks \citep{xu2022sparsenam}, generation of confidence intervals using spline basis expansion \citep{luber2023structnam}, and estimation of the skewness, heteroscedasticity, and kurtosis of the underlying data distributions \citep{thielmann2023namlss}.
Our work extends NAMs via Laplace-approximated Bayesian inference to endow them with principled uncertainty estimation and feature interaction and selection abilities within a unified Bayesian framework.

\paragraph{Bayesian neural networks.}

Bayesian neural networks promise to marry the expressivity of neural networks with the principled statistical properties of Bayesian inference \citep{mackay1992practical, neal1993bayesian}.
There are many approximate inference techniques, such as Laplace inference \citep{mackay1992practical, daxberger2021redux}, stochastic weight averaging \citep{maddox2019simple}, dropout \citep{gal2016dropout}, variational inference \citep{blundell2015bbb}, ensemble-based methods \citep{lakshminarayanan2017simple, wang2019function, d2021repulsive}, and sampling approaches \citep[e.g.,][]{neal1993bayesian, neal2011mcmc, welling2011bayesian}.
In our work, we leverage linearized Laplace~\citep{mackay1994bayesian,khan2019approximate,foong2019between,immer2021glm} for inference in Bayesian NAMs. 
This is motivated by the Laplace approximation's ability to provide reliable predictive uncertainties and marginal likelihood estimates, the latter of which are not readily available in most other approximations. 
Here, we devise a Laplace approximation that is specifically optimized for the structure of NAMs.

\paragraph{Additive Gaussian processes.}
Gaussian Processes (GPs) are a powerful and flexible class of probabilistic models. 
GPs are also compelling for the construction of GAMs as they are characterized by their ability to model complex relationships, providing uncertainty estimates along with their predictions. 
Additive GPs have been explored in studies by \citet{kaufman2010bayesian}, \citet{duvenaud2011additivegp}, \citet{timonen2021lgpr} and \citet{lu2022oak}.
Specifically, the Orthogonal Additive Kernel (OAK) proposed by \citet{lu2022oak} enables both selection of features and feature interaction of all orders relying on the efficient computational scheme of \citet{duvenaud2011additivegp} and the orthogonality constraint of \citet{durrande2012additive}.
Our work equips NAMs with similar features through Bayesian inference.
Refer to \cref{sec:detail_related_work} for a further detailed treatment of the related work.

\section{Laplace-Approximated NAM}

We introduce a Bayesian formulation of neural additive models and propose a tractable inference procedure, which is based on the linearized Laplace approximation of neural networks~\citep{mackay1991approx, khan2019approximate}.
The proposed model (\ourmethod) relies on a block-diagonal variant of the Gauss-Newton approximation with Kronecker factorization~\citep{martens2015kfac} and uses the associated predictive and log-marginal likelihood~\citep{immer2021scalable, immer2021glm} to estimate feature-wise uncertainty and automatically select features. 
Further, we identify promising feature interactions using mutual information within the posterior. 

\subsection{Bayesian Neural Additive Model}

We consider supervised learning tasks with inputs $\inputs_n \in \mathbb{R}^D$ and outputs $y_n$ where $y_n \in \mathbb{R}$ in regression and $y_n \in \{0,1\}$ in classification, and denote $\mathcal{D} = \{(\inputs_n, \targets_n)\}_{n=1}^N$ as the training set containing $N$ sample pairs.
A neural additive model is a neural network $f$ consisting of sub-networks $f_1, \dots, f_D$ with parameters $\allweights = \{\weights_1, \dots, \weights_D$\}, wherein each sub-network applies to an individual input dimension,
\begin{equation}
	f(\inputs;\,\allweights) = f_1(x_1;\,\weights_1) + \dots + f_D(x_D;\,\weights_D).
\end{equation}
We refer to the sub-networks $f_d: \mathbb{R} \times \mathbb{R}^P \to \mathbb{R}$, with parameters $\weights_d \in \mathbb{R}^P$, as \emph{feature networks}.
For simplicity, we assume that all feature networks share the same architecture.
In practice, it can vary to accommmodate mixed feature types such as binary or categorical features.
The sum is mapped to an output $y$ using a likelihood function $p(\mathcal{D}\,|\,\weights_1, \dots, \weights_D)$ with inverse link function $g^{-1}(\,\cdot\,)$, e.g., the sigmoid for classification, such that we have $\Esp[y\,|\,\inputs] = g^{-1}(f(\inputs))$, and
\begin{equation}
	p(\mathcal{D}\,|\,\allweights) 
		= \prod_{n=1}^N p(y_n\,|\,f(\inputs_n;\,\allweights)).
\end{equation}
We impose a zero-mean Gaussian prior distribution over the parameters of each feature network with prior precision hyperparameters $\precisions = \{\precision_1, \dots, \precision_D\}$,
\begin{equation}
    p(\allweights)
	= p(\weights_1, \dots, \weights_D\,|\,\precisions)
		= \prod_{d=1}^D 
			\mathcal{N}(\weights_d;\,\mathbf{0},\,\precision_d^{-1} \mathbf{I}).
\end{equation}
These terms adaptively regularize the network parameters and enable feature selection in a similar fashion to automatic relevance determination \citep[ARD;][]{mackay1994bayesian, neal1995bayesian}.
Large values push the corresponding feature networks toward zero and low values encourage highly non-linear fits.
In practice, one can also use separate priors and precision terms for each layer of each feature network, as this has been shown to be beneficial in linearized Laplace~\citep{immer2021scalable, antoran2022adapting}.

\subsection{Linearized Laplace Approximation}\label{sec:methlanam}

We devise a linearized Laplace approximation of the Bayesian NAM to obtain predictive uncertainties and select features as well as their interactions within a single framework, which we call \ourmethod.
The posterior predictive of the linearized Laplace is well-established and known to provide calibrated uncertainty estimates~\citep{daxberger2021redux}.
Further, its marginal likelihood estimates are useful to perform automatic relevance determination of feature networks during training and optimize observation noise parameters~\citep{mackay1994bayesian}.
Lastly, we use the mutual information in the posterior between feature network pairs to identify promising feature interactions.

We linearize the model around a parameter estimate $\allweights^*$,
\begin{align}
	f^{\mathrm{lin}}(\inputs;\,\allweights^*)
		&= f^{\mathrm{lin}}_1(x_1;\,\weights_1^*) 
		 + \dots 
		 + f^{\mathrm{lin}}_D(x_D;\,\weights_D^*), \\
	f^{\mathrm{lin}}_d(x_d;\,\weights_d^*) 
		&= f_d(x_d;\,\weights_d^*) 
		 + \mathcal{J}_{\weights^*}^{(d)}(x_d) (\weights_d - \weights_d^*),
\end{align}
where $\mathcal{J}_{\weights}^{(d)} : \mathbb{R} \to \mathbb{R}^{P}$ is the Jacobian of the $d$-th feature network w.r.t.~$\weights_d$, such that $[ \mathcal{J}_{\weights}^{(d)}]_i = \partial f_d / \partial \theta_{d,i}$.
This step follows from the linearity of the gradient operator and the fact that $\partial f_d / \partial \theta_{d'\!\!,\,i} = 0$ for $d \neq d'$.

This reduces the model to a generalized linear model in the Jacobians whose posterior can be approximated with a block-diagonal Laplace approximation as
\begin{equation}\label{eq:posterior}
	q(\weights) = \mathcal{N}(\weights^*,\,\boldsymbol{\Sigma}^*),
	\quad \boldsymbol{\Sigma}^*
	 	\approx \begin{bmatrix}
			\boldsymbol{\Sigma}_1 & \dots & 0 \\
			\vdots & \ddots & \vdots \\
			0 & \dots & \boldsymbol{\Sigma}_D \\
		\end{bmatrix}\!\!,
\end{equation}
where the diagonal covariance blocks are determined using the feature network Jacobians and second derivatives of the log-likelihood, $\gamma_n = -\frac{\partial^2}{\partial f^2}\log p(y_n\,|\,f_n)$, by
\begin{equation}
	\boldsymbol{\Sigma}_d =
		\left[
			\left[
				\sum_{n=1}^N 
					\gamma_n \cdot
					\mathcal{J}_{\weights^*}^{(d)}(x_{n,d})^{\top}
					\mathcal{J}_{\weights^*}^{(d)}(x_{n,d})
			\right]
			+ \precision_d \mathbf{I}
		\right]^{-1}\!\!\!\!\!\!\!\!.
\end{equation}
The approximation also leads to an additive structure over feature networks in the log-marginal likelihood,
\begin{align}
	\log p(\mathcal{D}\,|\,\precisions)
		&\approx \log p(\mathcal{D},\weights^*\,|\,\precisions) 
			-\tfrac{1}{2} \log |\tfrac{1}{2\pi} \boldsymbol{\Sigma}| \\
		&\geq \log p(\mathcal{D}, \weights^*\,|\,\precisions)
			-\tfrac{1}{2} 
				{\textstyle \sum_{d} \log |\tfrac{1}{2\pi} \boldsymbol{\Sigma}_d|}
		\label{equ:marglik}
\end{align}
where the lower bound is a consequence of the block-diagonal structure~\citep{immer2023ntk}.

In practice, we further approximate the covariance blocks $\boldsymbol{\Sigma}_d$ by using layer-wise Kronecker factorization \citep{ritter2018kfac, daxberger2021redux}, thereby avoiding the memory and computational complexity associated with the quadratic size in the number of parameters $P$.

\subsubsection{Feature Network Selection}

During training, the feature networks are implicitly compared and selected using adaptive regularization. 
This selection mechanism arises from the optimization of the lower bound on the Laplace approximation to the log-marginal likelihood in \cref{equ:marglik}, which we denote here as $\log q(\mathcal{D}\,|\,\precisions)$.
In the automatic relevance determination (ARD) procedure of \citet{mackay1994bayesian} and \citet{neal1995bayesian}, parameters of the first layer are grouped together according to their input feature and regularized as one.
In our case, one feature network corresponds to one group.

We maximize the lower bound on the log-marginal likelihood, $\max_{\precisions} \log q(\mathcal{D}\,|\,\precisions)$, by taking gradient-based updates\footnote{We also conducted experiments with the closed-form updates of \citet{mackay1991approx} and obtained comparable results.} during training with
\begin{equation}\label{eq:marglik}
	\tfrac{\partial}{\partial \precision_d} \log q(\mathcal{D}\,|\,\precisions)
		= \tfrac{P}{\precision_d} - ||\weights_d^*||_2^2 - \trace \boldsymbol{\Sigma}_d.
\end{equation}
An intuition of the corresponding closed-form update derived by \citet{mackay1991approx} is given in \citet{tipping2001sparse}:
the optimal value of $\precision_d$ is a measure of the concentration of $\boldsymbol{\Sigma}_d$ relative to the prior and depends on how well the data determines the parameters $\weights_d$.
Large values of $\precision_d$ lead to strong regularization and effectively switch off the $d$-th feature network.
This is depicted in \cref{fig:toyexample}, where our method demonstrates its ability to disregard an uninformative feature.
As a result, the procedure can lead to enhanced model interpretability and robustness as it redirects the attention to smaller subsets of informative features.

\subsubsection{Feature Network Predictive}

Due to linearization, we can obtain function-space predictive uncertainties in a closed form, like for Gaussian processes.
Given an unobserved sample $\inputs^*$, the predictive variance of the linearized model $f^{\mathrm{lin},\,*}$ corresponds to the sum of predictive variances of the linearized feature networks,
\begin{align}\label{equ:variancedecomp}
	\Var[f^{\mathrm{lin},\,*}\,|\,\inputs^*]
		&= {\textstyle \sum_d
			\Var[f_d^{\mathrm{lin},\,*}\,|\,x_d^*]
		} \\
		&= {\textstyle \sum_d 
			\mathcal{J}_{\theta^*}^{(d)}(x_d^*)^{\top}
			\boldsymbol{\Sigma}_d
			\mathcal{J}_{\theta^*}^{(d)}(x_d^*).
		}
\end{align}
This is due to the block-diagonal structure of our posterior approximation in \cref{eq:posterior}.
We discuss further properties and advantages of this in \cref{apd:indep}.

As training progresses, the feature networks may shift to satisfy a global intercept value in their sum. They should therefore be shifted back around zero before visualization by removing the expected contribution,
\begin{equation}\textstyle
	\hat{f}_d(x_d^*) = f_d(x_d^*) - \Esp_{\inputs \sim p(\mathcal{D})}[f_d(x_d)].
\end{equation}
Importantly, this adjustment does not affect the predictive variance $\Var[f_d^{\mathrm{lin},\,\ast}]$.
This variance estimate can be used to generate credible intervals for local and global explanations of the model, as shown in \cref{fig:mimiciiismall} and \ref{fig:locallanam}.
Notably, this allows the model to not only communicate on which data points it is uncertain, but also which features are responsible for the predictive uncertainty.
Also, refer to \cref{fig:toyexample} for an example of the obtained predictive uncertainties.

\begin{table*}[t]
  \caption[]{Negative test $\log$-likelihood (lower is better) on the UCI regression (top) and classification (bottom) benchmarks. Results better or within one standard error of the best performance of first-order methods in the first block are in bold. Results in {\color{OliveGreen} green} indicate an improvement beyond one standard error when compared to the corresponding model without second-order interactions. The \ourmethod performs competitively with the other additive models and often outperforms the \textsc{nam}. Moreover, when 10 interactions are selected, it almost always improves and often reaches competitive performance with the fully-interacting LightGBM.}
    \vspace{2mm}
    \begin{adjustbox}{max width=0.95	\textwidth,center}
        \begin{tabular}{l|rrr|rr|r}
 \toprule
Dataset & \textsc{nam} & \textsc{la-nam} & \textsc{oak-gp} & \textsc{la-nam}\textsubscript{10} & \textsc{oak-gp}\textsubscript{*} & LightGBM \\
 \midrule
autompg ($n$ = 392) & {\bfseries 2.69 ±0.16} & {\bfseries 2.46 ±0.08} & {\bfseries 2.55 ±0.10} & {\bfseries 2.45 ±0.09} & {\bfseries 2.46 ±0.14} & {\bfseries 2.53 ±0.07} \\
concrete ($n$ = 1030) & 3.46 ±0.12 & {\bfseries 3.25 ±0.03} & {\bfseries 3.19 ±0.09} & {\color{OliveGreen} {\bfseries 3.18 ±0.04}} & {\color{OliveGreen} {\bfseries 2.81 ±0.06}} & {\bfseries 3.09 ±0.09} \\
energy ($n$ = 768) & {\bfseries 1.48 ±0.02} & {\bfseries 1.44 ±0.02} & {\bfseries 1.46 ±0.02} & {\color{OliveGreen} {\bfseries 1.11 ±0.12}} & {\color{OliveGreen} {\bfseries 0.61 ±0.04}} & {\bfseries 0.81 ±0.05} \\
kin8nm ($n$ = 8192) & -0.18 ±0.01 & {\bfseries -0.20 ±0.00} & 0.09 ±0.01 & {\color{OliveGreen} {\bfseries -0.28 ±0.02}} & 0.09 ±0.01 & {\bfseries -0.50 ±0.03} \\
naval ($n$ = 11934) & -3.87 ±0.01 & -7.24 ±0.01 & {\bfseries -8.93 ±0.07} & {\color{OliveGreen} -7.44 ±0.07} & {\color{OliveGreen} {\bfseries -9.43 ±0.01}} & -5.19 ±0.01 \\
power ($n$ = 9568) & 2.89 ±0.02 & {\bfseries 2.85 ±0.01} & {\bfseries 2.81 ±0.03} & {\color{OliveGreen} {\bfseries 2.79 ±0.01}} & {\color{OliveGreen} {\bfseries 2.73 ±0.02}} & {\bfseries 2.67 ±0.02} \\
protein ($n$ = 45730) & 3.02 ±0.00 & 3.02 ±0.00 & {\bfseries 3.00 ±0.00} & {\color{OliveGreen} {\bfseries 2.94 ±0.01}} & {\color{OliveGreen} {\bfseries 2.88 ±0.00}} & {\bfseries 2.83 ±0.00} \\
wine ($n$ = 1599) & {\bfseries 1.02 ±0.04} & {\bfseries 0.98 ±0.03} & 1.14 ±0.04 & {\bfseries 0.97 ±0.03} & 1.67 ±0.65 & {\bfseries 0.96 ±0.03} \\
yacht ($n$ = 308) & 2.24 ±0.08 & {\bfseries 1.81 ±0.10} & {\bfseries 1.86 ±0.11} & {\color{OliveGreen} {\bfseries 0.76 ±0.20}} & {\color{OliveGreen} {\bfseries 0.79 ±0.17}} & {\bfseries 1.37 ±0.28} \\
\midrule
australian ($n$ = 690) & {\bfseries 0.38 ±0.04} & {\bfseries 0.34 ±0.03} & {\bfseries 0.33 ±0.03} & {\bfseries 0.34 ±0.03} & {\bfseries 0.35 ±0.03} & {\bfseries 0.31 ±0.03} \\
breast ($n$ = 569) & 0.16 ±0.03 & {\bfseries 0.10 ±0.02} & {\bfseries 0.07 ±0.01} & 0.10 ±0.02 & {\bfseries 0.09 ±0.01} & {\bfseries 0.09 ±0.01} \\
heart ($n$ = 270) & 0.41 ±0.04 & {\bfseries 0.33 ±0.02} & 0.42 ±0.06 & {\bfseries 0.33 ±0.03} & 0.42 ±0.06 & {\bfseries 0.39 ±0.04} \\
ionosphere ($n$ = 351) & 0.31 ±0.04 & {\bfseries 0.25 ±0.04} & {\bfseries 0.22 ±0.02} & {\bfseries 0.27 ±0.03} & {\bfseries 0.21 ±0.03} & {\bfseries 0.19 ±0.03} \\
parkinsons ($n$ = 195) & {\bfseries 0.29 ±0.04} & {\bfseries 0.26 ±0.03} & {\bfseries 0.27 ±0.03} & {\bfseries 0.25 ±0.03} & {\bfseries 0.21 ±0.02} & {\bfseries 0.22 ±0.03} \\
\bottomrule
\end{tabular}
    \end{adjustbox}
  \label{tab:uci10nll}
  \vspace{-2mm}
\end{table*}

\subsubsection{Feature Network Interaction}\label{sec:interaction}

Many datasets are not truly additive and thus require modeling interactions of features to adequately fit the data.
However, it is \emph{a priori} unclear which features exhibit underlying interactions.
As the search space grows exponentially for higher-order interactions, we focus here on the second order.
In second-order feature interaction detection, the goal is to find a subset of all existing $D^2$ interaction pairs that, when added to the model, maximize the gain in performance.
For each selected interacting pair $(d,\,d')$, we can then append a \emph{joint feature network} $f_{d,\,d'}(x_d,\,x_{d'})$ and perform fine-tuning of the model with the appended networks as part of a secondary training stage.

Our method for detecting and selecting second-order interactions makes use of the mutual information between feature networks.
If the mutual information between the feature network parameters $\weights_d$ and $\weights_{d'}$ is high, then conditioning on the values of either of these should provide information about the other and thus, their functions.
This can be an indication that a joint feature network for this pair may improve the data fit.

Although the mutual information between feature networks is zero in the approximation of \cref{eq:posterior}, this is not necessarily the case in the true posterior.
For the purpose of determining mutual information of the feature networks, we therefore fit separate last-layer Laplace approximations of the model for all feature pairs.
After the first training stage, we consider only the scalar output multiplier weights $\theta_d$ of each feature network and for each candidate pair of features $(d,\,d')$ determine the scalar marginal variances $\sigma_d^2$, $\sigma_{d'}^2$, and co-variance $\sigma_{d,d'}^2$ in the resulting $D \times D$ posterior covariance matrix, which is computationally efficient.
The mutual information can then be approximated using
\begin{align}
	\MutInf(\theta_d;\,\theta_{d'})
		&= \Ent(\theta_d) + \Ent(\theta_{d'}) - \Ent(\theta_d,\,\theta_{d'}) \\
		&\approx \tfrac{1}{2} \log \left[
			\sigma_d^2 \sigma_{d'}^2 (\sigma_d^2 \sigma_{d'}^2 - \sigma_{d,d'}^2)^{-1}
		\right] \\
		&= \tfrac{1}{2} \log \left[
			1 - \mathrm{Corr}(\theta_d,\,\theta_{d'})^2
		\right]^{-1}.
\end{align}
Finally, we select interactions by computing the mutual information for all feature pairs and taking the top-$k$ highest scoring pairs.
Importantly, this can be done after training in the first stage and without the need for an additional model to assess interaction strength.
We show in the experiments that with only a few second-order interactions selected this way the performance of the additive model can improve to be at the level of a fully-interacting baseline.

\section{Experiments}
\label{sec:experiment}

We evaluate the proposed \ourmethod on a collection of synthetic and real-world datasets emphasizing its potential for supporting decision-making in the medical field.\footnote{Code: \href{https://github.com/fortuinlab/LA-NAM}{\texttt{github.com/fortuinlab/LA-NAM}}}
To assess performance, we compare against the original \textsc{nam} of \citet{agarwal2021nams} and other generalized additive models, namely, a smoothing spline \emph{generalized additive model} (\textsc{gam}), with smoothing parameters selected via cross-validation \citep{hastie1999gams,serven2018pygam}, and a GP with an \emph{orthogonal additive kernel} \citep[\textsc{oak-gp};][]{lu2022oak}.
We also include LightGBM \citep{ke2017lightgbm} as a state-of-the-art fully-interacting model which approximates the maximal attainable performance in tabular regression and classification tasks \citep{grinsztajn2022tree}.
Detailed information regarding the experimental setup is provided in \cref{apd:expdetails}.

\begin{table*}[t]
\caption[]{Comparison of model performance on ICU mortality prediction tasks. Mean and standard error reported across five seeds. AUROC and AUPRC are in percent. Bold indicates best performance first-order methods; {\color{OliveGreen} green}/{\color{BrickRed} red} highlights an improvement/decrease with interactions. The \textsc{oak-gp} runs out of memory for HiRID due to the number of features. The \ourmethod generally outperforms the \textsc{nam}, and improves with 10 interacting features on MIMIC-III, becoming competitive with the fully-interacting LightGBM gold-standard.\label{tab:medical}}
    \vspace{2mm}
    \begin{adjustbox}{max width=0.95\textwidth,center}
        \begin{tabular}{lccccccccc}
\toprule
 Task & \multicolumn{3}{c}{MIMIC-III ICU Mortality} & \multicolumn{3}{c}{HiRID ICU Mortality} \\
 \cmidrule(lr){2-4} \cmidrule(lr){5-7} \cmidrule(lr){8-10}
 Metric & AUROC ($\uparrow$) & AUPRC ($\uparrow$) & NLL ($\downarrow$) & AUROC ($\uparrow$) & AUPRC ($\uparrow$) & NLL ($\downarrow$) \\
\midrule
\textsc{nam} & 77.6 ± 0.03 & 32.3 ± 0.03 & 0.274 ± 8e-5 & 89.6 ± 0.17 & {\bfseries 60.7 ± 0.14} & 0.228 ± 1e-2 \\
\textsc{la-nam} & 79.6 ± 0.01 & 34.8 ± 0.04 & 0.264 ± 5e-5 & {\bfseries 90.1 ± 0.03} & {\bfseries 60.5 ± 0.14} & {\bfseries 0.174 ± 2e-4} \\
\textsc{oak-gp} & {\bfseries 79.9 ± 0.03} & {\bfseries 35.2 ± 0.11} & {\bfseries 0.263 ± 1e-4} & --- & --- & --- \\
\midrule
\textsc{la-nam}\textsubscript{10} & {\color{OliveGreen} {\bfseries 80.2 ± 0.10}} & {\color{OliveGreen} {\bfseries 35.2 ± 0.06}} & {\color{OliveGreen} {\bfseries 0.262 ± 3e-4}} & {\bfseries 90.1 ± 0.01} & {\bfseries 60.5 ± 0.20} & {\bfseries 0.174 ± 4e-4} \\
\textsc{oak-gp}\textsubscript{*} & {\color{BrickRed}71.7 ± 0.66} & {\color{BrickRed}28.5 ± 0.42} & {\color{BrickRed}0.288 ± 1e-3} & --- & ---  & --- \\
\midrule
LightGBM & {\bfseries 80.6 ± 0.08} & {\bfseries 35.6 ± 0.19} & {\bfseries 0.261 ± 3e-4} & {\bfseries 90.7 ± 0.00} & {\bfseries 61.6 ± 0.00} & {\bfseries 0.172 ± 0.00} \\
\bottomrule
\end{tabular}

    \end{adjustbox}
    \vspace{-2mm}
\end{table*}

\begin{figure*}[t]
    \makebox[\textwidth]{
        \includegraphics[width=\textwidth]{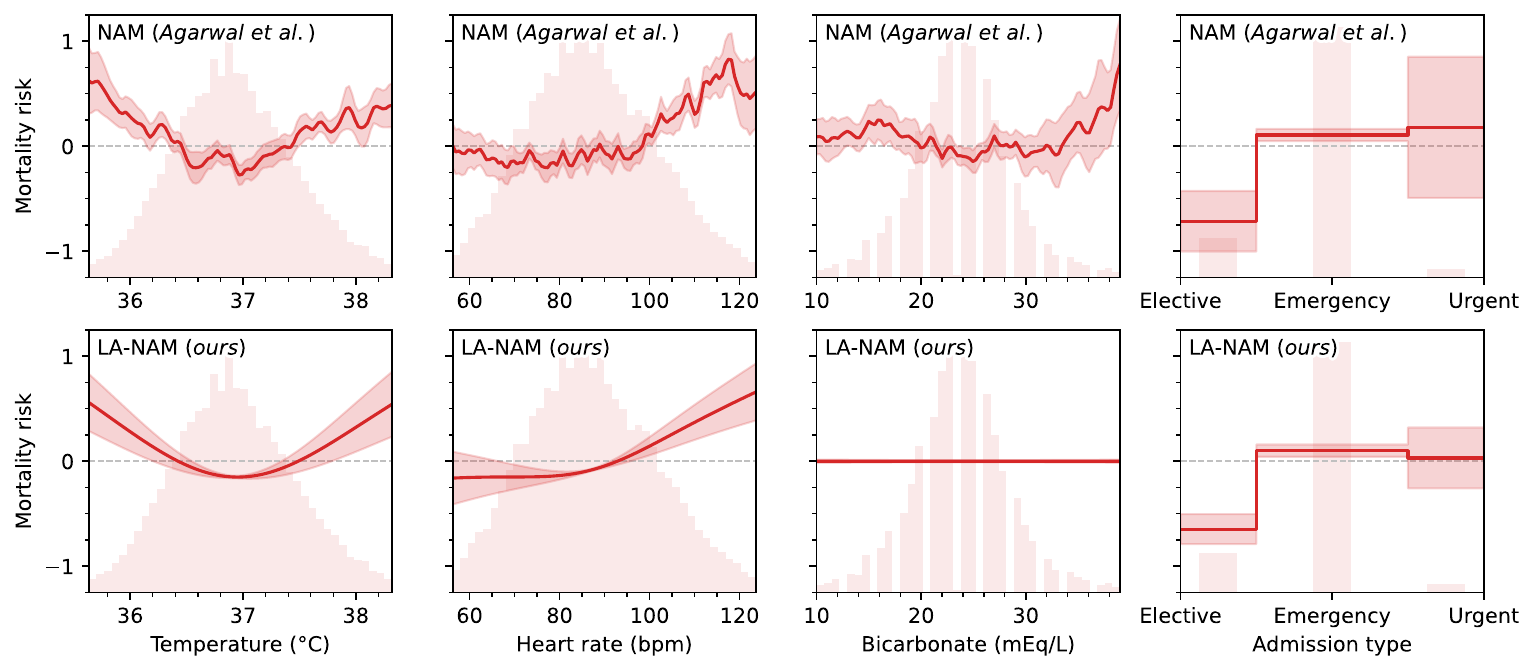}
    }
    \caption[]{Risk of mortality and associated epistemic uncertainty ($\pm$ 2 std.~deviations) on the MIMIC-III mortality prediction task. The \ourmethod relies on smoother feature curves and provides useful uncertainties while ignoring the uninformative feature.\label{fig:mimiciiismall}}
\end{figure*}

The \textsc{nam} lacks a built-in notion of epistemic uncertainty.
As such, figures which provide the uncertainty associated with its predictions use the standard deviation of the recovered functions across the deep ensemble of feature networks.

\subsection{Illustrative Example}
\label{sec:exptoyexample}

To demonstrate the capability of recovering purely additive structures from noisy data, we constructed a synthetic regression dataset for which the true additive functions are known.
Generalized additive models are expected to accurately recover the additive functions present in this dataset since it is designed in such a way that there is no interaction between the input features.

In \cref{fig:toyexample}, we show the recovered functions for $f_1$ and $f_4$ along with the ground-truth quadratic function $f_1$ and zero function $f_4$. 
The \textsc{nam} exhibits pronounced jumpy behavior due to the presence of noise, resulting in a considerably poor mean fit.
In contrast, the proposed \ourmethod fits the data accurately while maintaining a good estimate of epistemic uncertainty.
It is less susceptible to misattributing noise to the recovered functions compared to the \textsc{nam}.
This is particularly striking for the uninformative function $f_4$, since only the \ourmethod and \textsc{oak-gp} correctly identify that it should have no effect, due to their Bayesian model selection.
More details and visualizations are in \cref{apd:toyexample}.

\begin{figure*}
    \makebox[\textwidth]{
        \includegraphics[width=0.8\textwidth]{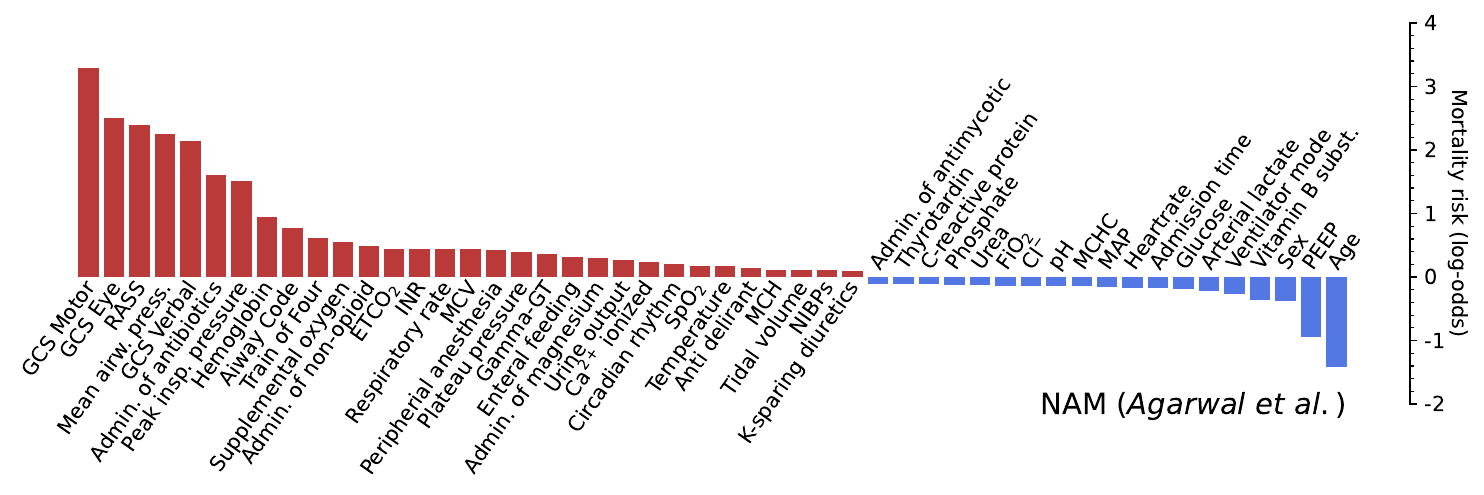}
    }
    \centering
    \caption[]{Local explanations for the risk of a sample patient in the HiRID mortality task in the \textsc{nam}. Features which contribute less than 0.1 to the log-odds magnitude are omitted. The \textsc{NAM} selects a large number of features and does not provide uncertainty estimates.}
    \label{fig:localnam}
\end{figure*}

\begin{figure}
	\includegraphics[width=\linewidth]{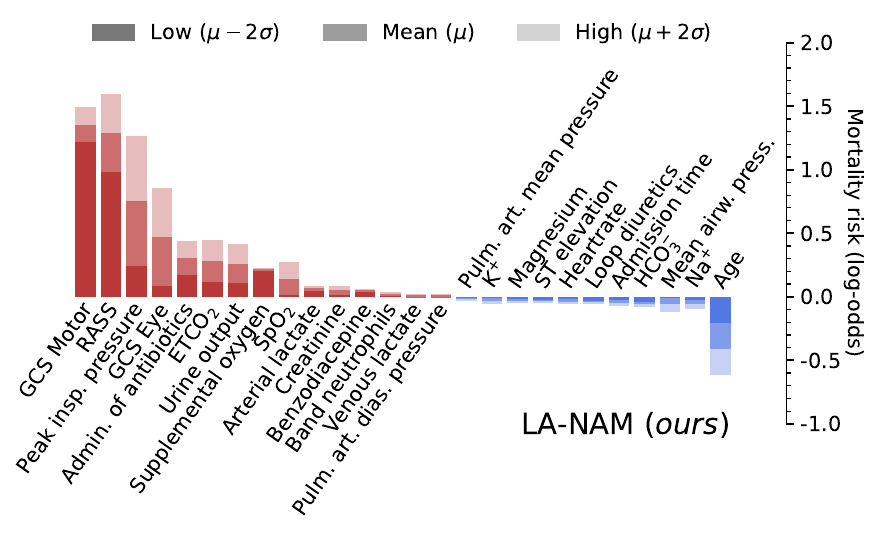}
    \caption[]{Local explanations from the \ourmethod for the same patient as in \cref{fig:localnam}. Features whose credible intervals overlap with zero are omitted. The model selects far fewer features and provides uncertainty estimates, further aiding interpretation.}
    \label{fig:locallanam}
\end{figure}

\subsection{UCI Regression and Classification}
\label{sec:expucibench}

We benchmark the \ourmethod and baselines on the standard selection of UCI regression and binary classification datasets.
Each dataset is split into five cross-validation folds and the mean and standard error of model performance are reported across folds.
We split off 12.5\% of the training data as validation data for the \textsc{nam}.
This extra validation data is not required for the \ourmethod, since it is tuned using the estimated $\log$-marginal likelihood.
Additionally, both the \ourmethod and \textsc{oak-gp} have support for selecting and fitting second-order feature interactions, further enhancing their modeling capacity.
For these models, we also present results where we have enabled feature interactions: We identify and fine-tune the top-10 feature pairs of the \ourmethod (which we denote as \textsc{la-nam}\textsubscript{10}), and increase the \textsc{oak-gp}'s maximum interaction depth to 2 so that it models \emph{all} pairwise interactions (denoted as \textsc{oak-gp}\textsubscript{*}).

In \cref{tab:uci10nll}, we report the negative $\log$-likelihood averaged over test samples.
The \textsc{nam} does not provide an estimate of the observation noise in regression, so it is assigned a maximum likelihood fit using its training data.
The \ourmethod consistently demonstrates competitive performance across multiple datasets. 
It tends to exhibit lower average negative $\log$-likelihood, indicating better performance, compared to the \textsc{nam} and performs comparatively well versus the \textsc{oak-gp}.
\textsc{la-nam}\textsubscript{10}, which refers to the fined-tuned model with top-10 feature interactions, almost always improves performance on regression when compared to its non-feature-interactive counterpart and often reaches the performance of the fully-interacting LightGBM.
Note that this might also be related to its ability to ignore uninformative features, which has been identified as a main weakness of neural networks on tabular data compared to tree-based methods \citep{grinsztajn2022tree}. A wider set of models along with additional performance and calibration metrics are considered in \cref{apd:uciresults,apd:ucicalib}.
\begin{figure*}
\begin{minipage}{.45\textwidth}
  \centering
  \includegraphics[width=\textwidth]{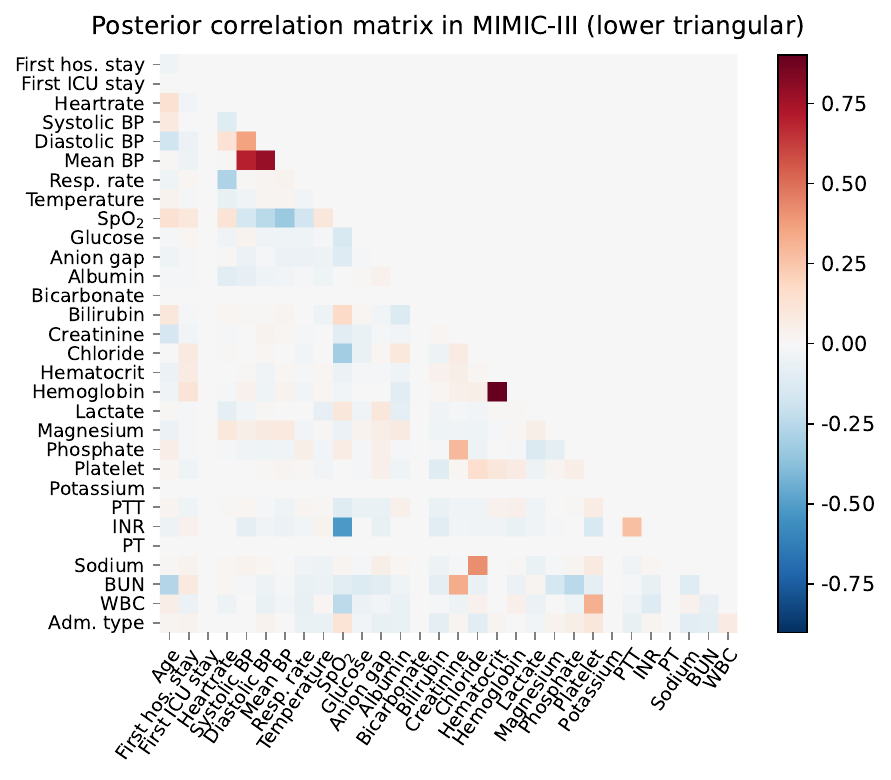}
\end{minipage}%
\hspace{1em}
\begin{minipage}{.55\textwidth}
  \centering
  \includegraphics[width=\textwidth]{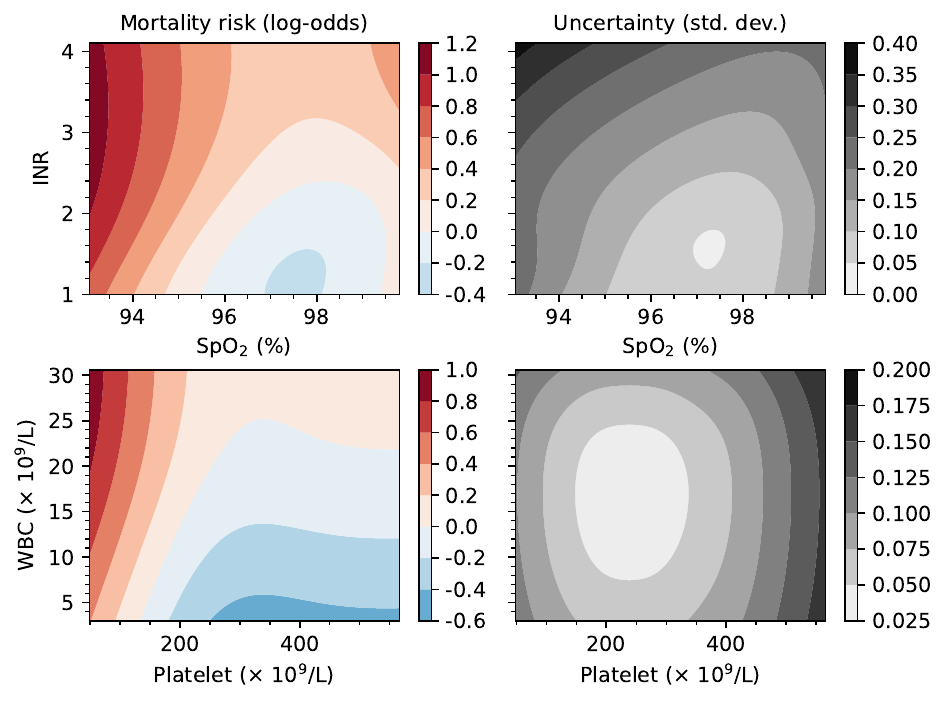}
\end{minipage}
\caption[]{Feature interactions uncovered in the MIMIC-III dataset by the \ourmethod. (left) Last-layer posterior correlation matrix, used to select the most informative feature-interaction pairs. (right) Two selected example feature interactions and their associated uncertainty.}
\label{fig:mimiciiiinteraction}
\end{figure*}

\subsection{Intensive Care Unit Mortality Prediction}\label{sec:expmimiciii}

To gain insights into the behavior of our method within a real-world clinical context, we investigate the performance in predicting patient mortality based on vital signs recorded 24 hours after admission into an intensive care unit (ICU).
To accomplish this, we utilize the MIMIC-III patient database \citep{johnson2016mimiciii} and employ the pre-processing outlined by \citet{lengerich2022mimiciii}. Additionally, we leverage the HiRID database \citep{faltys2021hirid} and adopt the pre-processing proposed by \citet{yeche2021hiridicubenchmark}.
Notably, our objective extends beyond achieving competitive predictive performance: we aim to provide valuable insights into the underlying sources of risk within the ICU to facilitate a clearer understanding of critical care dynamics.

\paragraph{Predictive performance.}

\cref{tab:medical} presents a summary of the test performance achieved by the various methods.
For the MIMIC-III dataset, a total of 14,960 patients were considered, with 2,231 held out as the test set.
Similarly, the HiRID dataset comprises 27,347 patients, with 8,189 held out for testing purposes.
On these tasks, the \ourmethod demonstrates superior performance compared to the \textsc{nam} in the key evaluation metrics of area under the ROC and precision-recall curve (AUROC and AUPRC), as well as average negative 
$\log$-likelihood (NLL).\footnote{Calibration is also improved, as shown in \cref{apd:medcalib}.}
\cref{fig:mimiciiismall} showcases a subset of the recovered additive structure from the MIMIC-III dataset.\footnote{Complete visualization is provided in \cref{apd:mimiciii}.}
In each subplot, the background displays a histogram depicting the distribution of feature values.
Note that the \textsc{nam} exhibits the same jumpy behavior observed in the synthetic example, adversely affecting interpretability, while the \ourmethod yields smoother curves, due to its optimized Bayesian prior. We find that the identified relationships of the displayed variables and risk by \ourmethod appear consistent with medical knowledge (personal communication with medical experts).
We also find that the \ourmethod effectively captures and quantifies epistemic uncertainty, aligning with the presence or absence of sufficient samples.

\paragraph{Feature selection.}

This experiment also illustrates the ability of the \ourmethod to assess the relevance of features through the selection of feature networks.
Because of their linear dependency, both high bicarbonate levels and low anion gap are indicators of metabolic acidosis~\citep{kraut2010acidosis1}.
The \ourmethod determines that the risk associated with bicarbonate can be adequately captured using only the measurement of the anion gap. Consequently, as demonstrated in \cref{fig:mimiciiismall}, it entirely disregards the bicarbonate feature.
See \cref{apd:ablation} for an ablation experiment confirming this.

\paragraph{Interpretability.}

The epistemic uncertainty and feature-selecting property of the \ourmethod play significant roles in the effectiveness of the generated local explanations.
This advantage is clear in \cref{fig:localnam,fig:locallanam}, where we compare the breakdown of risk factors between the \textsc{nam} and \ourmethod, respectively. %
We show that the \ourmethod selects a noticeably condensed, more concise set of features.
Moreover, it provides uncertainties in local explanations, bringing valuable insight by acknowledging the inherent variability and potential limitations of the model.
In this case, it allows clinicians to gauge the reliability and robustness of the predicted risk factors, enabling more informed decision-making and facilitating trust in the model's outputs.

\paragraph{Feature interactions.}

In addition to the first-order models, both the \textsc{la-nam} and \textsc{oak-gp} were evaluated with feature interactions (\textsc{la-nam}\textsubscript{10} and \textsc{oak-gp}\textsubscript{*} in \cref{tab:medical}).
Incorporating these interactions led to significantly improved performance for the \textsc{la-nam}\textsubscript{10} on the MIMIC-III dataset, reaching competitive performance with the fully-interacting LightGBM gold standard.
On the left side of \cref{fig:mimiciiiinteraction}, the last-layer correlation matrix for second-order interactions is presented, revealing the relationships between the different feature pair candidates for inclusion in the second-order fine-tuning stage. On the right side, specific interactions involving the risk factors of WBC and Platelet, as well as INR and SpO\textsubscript{2}, are depicted. In particular, we show an increase in mortality risk for high white blood cell and low platelet counts as well as for elevated INR and low oxygen saturation. While the first interaction is characteristic of immune thrombocytopenia~\citep{cooper2019immune}, a known risk factor for critically ill patients~\citep{baughman1993thrombocytopenia,trehel2012clinical}, the second has not been studied in the literature. Hence, by enabling second-order interaction selection, \ourmethod may also help to discover new risk factors (of course, follow-up studies would be needed).

\section{Conclusion}
\label{sec:conclusion}

In this work, we have introduced a Bayesian adaptation of the widely used neural additive model and derived a customized linearized Laplace approximation for inference. This approach allows for a natural decomposition of epistemic uncertainty across additive subnetworks, as well as implicit feature selection by optimizing the $\log$-marginal likelihood.
Our empirical results illustrate the robustness of our Laplace-approximated neural additive model (\ourmethod) against noise and uninformative features. Furthermore, when allowed to autonomously select its feature interactions, the \ourmethod demonstrates performance on par with fully interacting gold-standard baselines.
The \ourmethod thus emerges as a viable option for applications with safety considerations and as a tool for data-driven scientific discovery.
Moreover, our work underscores the potential for future research at the intersection of interpretable machine learning and Bayesian inference.
Overall, this work contributes to the vision of powerful, robust, yet transparent and understandable machine learning models. We hope that our work will inspire further advances in Bayesian additive models, marrying transparency with probabilistic modeling.

\subsection*{Impact Statement}

Since this is basic research, we do not foresee any immediate broader impact.
However, as the proposed model can lend itself well to medical tasks (see \cref{sec:expmimiciii}), it could have an impact on improving medical care.
These features may also have relevance in other critical domains where the use of non-transparent models can be problematic, such as jurisdiction, hiring, or finance.

Additionally, in applications where fairness and implicit biases are important, detecting biases or potential lack of fairness could be facilitated by the feature-independence property and the feature-selecting capabilities of our model.
It should be noted that existing biases in the data will likely propagate to the model, so this should be critically assessed.

\subsection*{Acknowledgments}

We wish to thank Ben Lengerich for providing us with the pre-processed MIMIC-III dataset used in the paper.

This project was supported by grant \#2022-278 of the Strategic Focus Area ``Personalized Health and Related Technologies (PHRT)'' of the ETH Domain.
A.\,I.~acknowledges funding by the Max Planck ETH Center for Learning Systems (CLS).
V.\,F.~was supported by a Branco Weiss Fellowship.

\bibliography{paper}
\bibliographystyle{icml2024}

\newpage
\appendix
\onecolumn

\renewcommand\thefigure{\thesection\arabic{figure}}
\setcounter{figure}{0}
\renewcommand\theequation{\thesection\arabic{equation}}
\setcounter{equation}{0}
\renewcommand\thetable{\thesection\arabic{table}}
\setcounter{table}{0}

\makeatletter
\renewcommand*\l@section{\@dottedtocline{1}{1.5em}{2.3em}}
\makeatother
\setcounter{page}{1}
\counterwithin{figure}{section}
\counterwithin{equation}{section}
\counterwithin{table}{section}

\section{Additional Theoretical Results and Discussion}

In this section, we provide additional details on the derivation of our approximate posterior for Bayesian NAMs and discuss in further detail the interaction detection procedure and theoretical computational bounds.

\subsection{Feature Network Independence}\label{apd:indep}

The approximate posterior defined in \cref{eq:posterior} results in feature networks that are mutually independent due to the block-diagonal structure of the covariance matrix.
This independence is needed for the decomposition of the predictive variance in \cref{equ:variancedecomp}.
We elaborate here on the motivation behind this independence assumption.

\begin{figure}[h]
    \centerline{
        \includegraphics[width=0.6\textwidth]{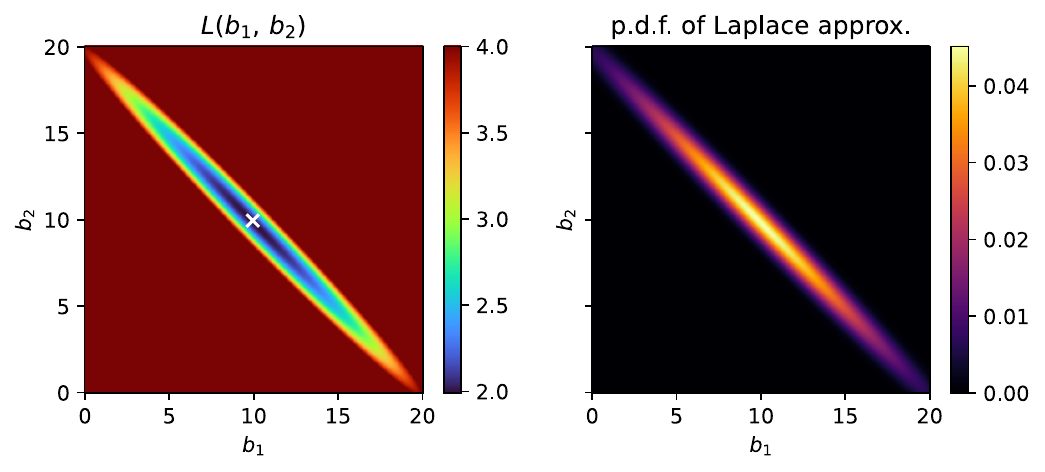}
    }
    \caption[]{Laplace approximation of the illustrative example in \cref{apd:indep}. ($C = 20$, $\lambda = 0.01$)}
    \label{fig:biascovar}
\end{figure}

As a thought experiment, suppose we wanted to find estimates of two variable terms $b_1$ and $b_2$, such that their sum is equal to some constant $C$.
We also desire that neither term $b_1$ or $b_2$ dominate the other so that they are roughly equally balanced.
One possible setup for finding a \emph{maximum a posteriori} (MAP) estimate of $b_1$ and $b_2$ could be to design a cost function $L(b_1,\,b_2)$ where the MAP solution is a minimizer,
\begin{equation}
    p(b_1,\,b_2) = \mathcal{N}(
        [b_1,\,b_2]\trans;\,\mathbf{0},\,\lambda^{-1} \mathbf{I}),
    \quad
    p(C\,|\,b_1,\,b_2) = \mathcal{N}(C;\,b_1 + b_2,\, 1),
\end{equation}
\vspace{-17.05pt}
\begin{align}
    \log p(b_1,\,b_2\,|\,C) 
        &\propto \log p(C\,|\,b_1,\,b_2) + \log p(b_1,\,b_2) \\
        &\propto -(b_1 + b_2 - C)^2 - \lambda (b_1^2 + b_2^2)
            \quad \eqdef -L(b_1,\,b_2).
\end{align}
For illustrative purposes, we set $C = 20$ and $\lambda = 0.01$.
In the left side of \cref{fig:biascovar}, we visualize the cost function values $L(b_1,\,b_2)$ and highlight the MAP solution as a white cross.
Now, suppose we are interested in finding an approximate posterior distribution for $b_1$ and $b_2$ with a Laplace approximation centered at the MAP estimate.
When we treat $b_1$ and $b_2$ as jointly dependent variables, we obtain the Gaussian approximate posterior distribution depicted on the right side of \cref{fig:biascovar}.

\noindent In this approximation, $b_1$ and $b_2$ exhibit strong anti-correlation.
This can be attributed to the observation that adjusting $b_1$ by some amount $\Delta$ can be accounted for by adjusting $b_2$ by $-\Delta$, since $b_1 + b_2 = (b_1 + \Delta) + (b_2 - \Delta)$.
Within the realm of Bayesian NAMs, this property is considered undesirable.
Ideally, we want the credible intervals for a feature network $f_d$ to solely reflect changes in its shape, without being influenced by vertical translations of other feature networks.
To circumvent the scenario illustrated above, one can adopt a strategy where $b_1$ and $b_2$ are treated as independent variables.
This involves performing a first Laplace approximation for $b_1$ while keeping $b_2$ fixed, followed by a separate Laplace approximation for $b_2$ keeping $b_1$ fixed.
In our model, this is ensured by performing separate Laplace approximations for each feature network.

Nonetheless, the true posterior may contain strong statistical dependency across the feature networks.
Therefore, for feature pairs which demonstrate high mutual information (\cref{apd:intermi}) or exhibit significant improvement on the marginal likelihood bound (\cref{apd:intermodsel}), we suggest to introduce second-order joint feature networks.
The purpose of these networks is to explicitly capture these dependencies in the resulting fine-tuned model.

\subsection{Feature Interaction Selection}\label{app:interaction}
In this section, we expand on the feature interaction selection based on the mutual information and also provide an alternative selection procedure which is phrased as model selection and optimization of the marginal likelihood.

\subsubsection{Feature Interaction via Mutual Information}\label{apd:intermi}
To select feature interactions, we consider the mutual information (MI) between the posteriors of feature networks.
If the mutual information between $\vtheta_d$ and $\vtheta_{d'}$ is high, it is simultaneously high between the functional posteriors $f_d$ and $f_{d'}$.
This means that conditioning on the output of feature network $f_d$ can provide information about the posterior of feature network $f_{d'}$, and therefore be an indication that a joint feature network for the interacting features could ultimately improve the \ourmethod fit.
In a separate dense covariance matrix Laplace approximation, each candidate pair $(d, d')$ has two marginal posteriors and a joint posterior.
These are Gaussian distributions with corresponding covariance matrices $\mSigma_{d}, \mSigma_{d'}$, and $\mSigma_{d,\,d'}$.
The mutual information can be approximated using
\begin{align}
\label{app:eq:mi_param}
    \MutInf(\vtheta_d;\,\vtheta_{d'}) &\approx - \half \log |\mSigma_{d,\,d'} (\mSigma_{d} \oplus \mSigma_{d'})\inv |
    = -\half \log [|\mSigma_{d,\,d'}| |\mSigma_{d}\inv| |\mSigma_{d'}\inv|],
\end{align}
where $\oplus$ denotes block-diagonal concatenation.
Constructing the full covariance matrix can be prohibitively expensive for a \ourmethod containing many feature networks.
Instead, we can iteratively compute the covariance blocks $\boldsymbol{\Sigma}_{d,\,d'}$ and determine the mutual information of each pair using Jacobians and precision matrices which are at most $2P \times 2P$ large \citep{daxberger2021subnet}.
Alternatively, we propose to use a last-layer approximation in \cref{sec:interaction} which turns the computation into $D \times D$ matrices and relies on scalar marginal variances.

For linearized Laplace approximations one can show that this is closely related to maximizing the mutual information of the posterior on $f_d$ and $f_{d'}$ over the training set.
In other words, how much information is gained about $f_d$ when observing $f_{d'}$ and vice-versa.
The joint posterior predictive on $f_d$ and $f_{d'}$ is a Gaussian with a $2 \times 2$ covariance matrix per data point, meaning $N \times N$ marginal predictive covariance matrices $\mS_{d}$ and $\mS_{d'}$, and a $2N\times 2N$ joint predictive covariance matrix $\mS_{d,\,d'}$, when considering the entire training set.
In similar fashion to \cref{app:eq:mi_param}, the mutual information of the functional posterior distribution can be determined by taking 
\begin{align}
\label{app:eq:mi_func}
    \MutInf(\vf_d;\,\vf_{d'}) = -\half \log |\mS_{d,\,d'} (\mS_{d} \oplus \mS_{d'})\inv|
    = -\half \log [|\mS_{d,\,d'}| |\mS_{d}\inv |\mS_{d'}\inv|].
\end{align}
Moreover, taking $\mJ_{d,\,d'} \in \R^{2N \times 2P}$ as the Jacobian matrix of $[f_d,\,f_{d'}]\transpose$ with respect to $[\boldsymbol{\theta}_d,\,\boldsymbol{\theta}_{d'}]\transpose$ over the training data, with brackets denoting concatenation, we can write the marginal and joint predictive covariances as
\begin{equation}\label{app:eq:mi_expand}
    \mS_{d,\,d'} = \mJ_{d,\,d'} \mSigma_{d,\,d'} \mJ_{d,\,d'}\transpose
    \quad \mathrm{and} \quad
    \mS_{d} \oplus \mS_{d'} = \mJ_{d,\,d'} (\mSigma_{d} \oplus \mSigma_{d'}) \mJ_{d,\,d'}\transpose.
\end{equation}
Plugging \cref{app:eq:mi_expand} back into the mutual information in \cref{app:eq:mi_func}, we see that it is equivalent to the parametric mutual information in \cref{app:eq:mi_param} when we have $N = P$ and full-rank Jacobians.
This shows that the parametric mutual information can also be thought of as an approximation to the functional mutual information.

\subsubsection{Feature Interaction as Model Selection}\label{apd:intermodsel}
Alternatively, we can phrase the selection of feature interactions as model selection, thereby optimizing the marginal likelihood of the Bayesian model.
Mathematically, we can achieve this by choosing the feature pairs that maximize the lower bound to the $\log$-marginal likelihood which stems from the factorized block-diagonal posterior.
Adding any feature interaction in the posterior will reduce the slack of the bound in \cref{eq:marglik} and we aim to choose the ones reducing it the most.
For any feature pair $(d,\,d')$, taking $\mathbf{J}_i \in \mathbb{R}^{N \times P}$ as the Jacobian matrix of $f_i$ with respect to $\boldsymbol{\theta}_i$ over training data, we can estimate the improvement on the bound by considering
\begin{align}
    \textrm{gain}(d,\,d') &= \sum_{i \in \{d,\,d'\}}\!\!\! [\log |\underbrace{\mJ_i\transpose \diag[\boldsymbol{\gamma}] \mJ_i + \lambda_i \mI}_{P \times P}|] - \log |\underbrace{\mJ_{d,\,d'}\transpose \diag[\boldsymbol{\gamma}] \mJ_{d,\,d'} + \lambda_d \mI \oplus \lambda_{d'} \mI}_{2P \times 2P}| \\
    &= \log |\mP_{d}| + \log |\mP_{d'}| - \log |\mP_{d,\,d'}|
    = - \log [|\mP_{d,\,d'}| |\mP_{d}\inv| |\mP_{d'}\inv|],
\label{eq:interaction_gain}
\end{align}
where $\diag[\boldsymbol{\gamma}]$ denotes a diagonal matrix containing $\gamma_1, \dots, \gamma_N$.
This quantifies the extent to which the log-marginal likelihood of the \ourmethod is enhanced when accounting for the posterior interaction between feature networks $f_d$ and $f_{d'}$.
The gain heuristic is derived from subtracting the approximate log-marginal likelihood of the initial model from one where a joint feature network is added.

As with the mutual information procedure, one can simplify this into a scalar case, thus maximizing normalized joint precision values instead of correlations.
This makes intuitive sense as well since the precision values can indicate pairwise independence when conditioning on all other variables, that is for all $d^{\prime\prime} \notin \{d,\, d'\}$.

\subsection{Computational Considerations and Complexities}

We briefly discuss the complexity of the proposed method, taking both computation and storage of the various quantities in consideration.
For simplicity, we assume a \ourmethod with $D$ feature networks, each with fixed number of parameters $P$, even though in practice the number of parameters can vary among feature networks.

In order to compute the block-diagonal approximate posterior of \cref{eq:posterior} we must determine and invert a Laplace-\textsc{ggn} posterior precision matrix for each feature network, resulting in total complexity of $\mathcal{O}(DNP^2)$ and $\mathcal{O}(DP^3)$, respectively.
Storage of the matrix can be done in $\mathcal{O}(DP^2)$ and computing its determinant in $\mathcal{O}(DP^3)$.
This can be overly prohibitive for large numbers of parameters, but can be significantly alleviated by using a layer-wise Kronecker-factored approximation \citep[\textsc{kfac}-\textsc{ggn};][]{martens2015kfac}.
Assuming an architecture with a single layer with hidden size $\mathcal{O}(\sqrt{P})$, \textsc{kfac}-\textsc{ggn} reduces computation of the precision matrix to $\mathcal{O}(DNP)$ and its inversion to $\mathcal{O}(DP^{3/2})$.
Further details are provided in \citet{immer2021glm}.

A separate Laplace approximation is required for both of the proposed feature interaction selection methods.
The approximate covariance matrix must contain off-diagonal blocks since both the mutual information and improvement on the log-marginal likelihood depend on joint covariance matrices for each candidate pair.
One option is to perform sub-network Laplace inference \citep{daxberger2021subnet} in order to iteratively score the pairs, which equates to taking $D\cdot(D-1)\,/\,2$ separate Laplace approximations over $2P$ parameters.
We propose instead to perform a last-layer approximation by only considering the $D$ output weights of each feature network, which results in computation in $\mathcal{O}(ND^2)$, storage in $\mathcal{O}(D^2)$, and inversion for mutual information in $\mathcal{O}(D^3)$.

\section{Additional Results and Experiments}

\subsection{Comparison of Approximate Inference Methods}
\label{apd:approxinf}

\begin{figure}[!h]
	\includegraphics[width=\textwidth]{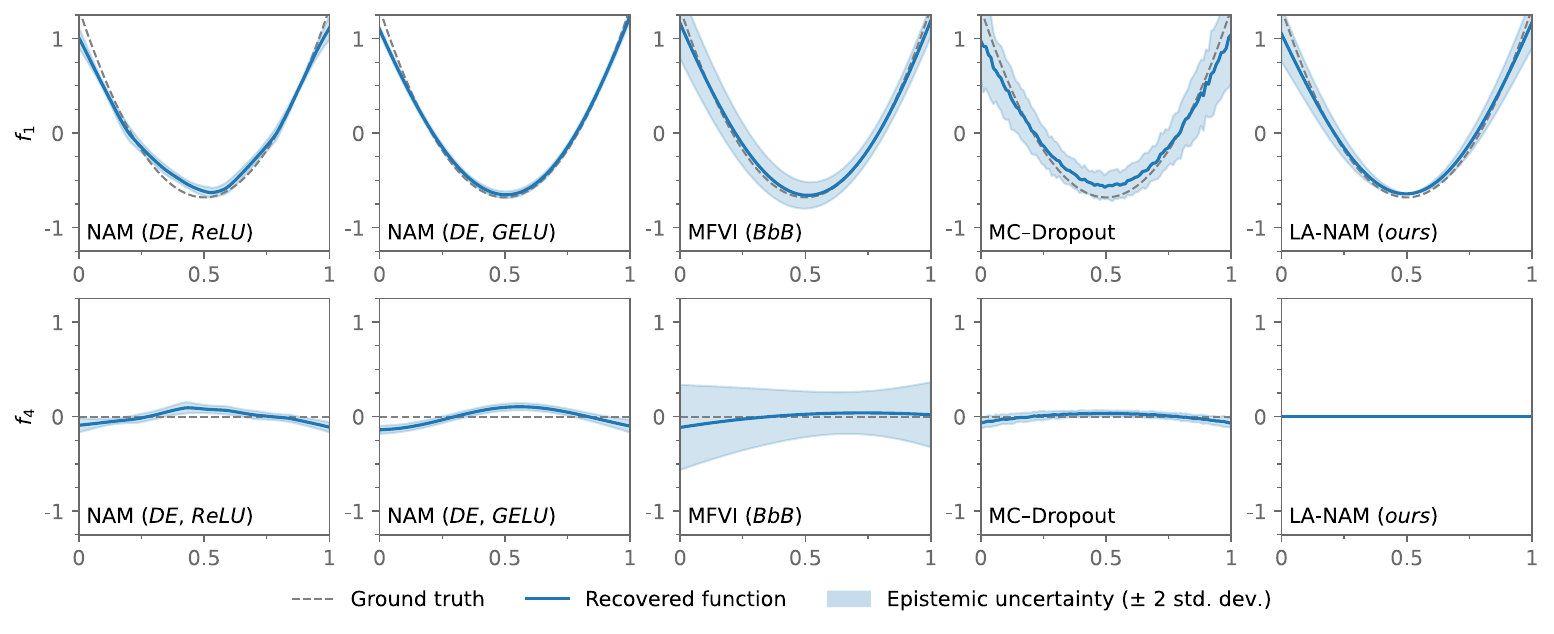}
	\centering
	\caption[]{Comparison of approximate inference methods for Bayesian NAMs on the synthetic data of \cref{sec:exptoyexample}. All feature networks use one layer of 64 GELU units with the exception of the leftmost model, ``NAM (\emph{DE, ReLU})'', which uses 64-64-32 ReLU units.}
	\label{fig:altapproxinf}
\end{figure}

We provide a brief overview of alternative approximate inference methods which can be employed to quantify uncertainty of feature networks in Bayesian NAMs.
\cref{fig:altapproxinf} displays the recovered functions and predictive intervals generated using these methods on the toy example of \cref{sec:exptoyexample}.

\paragraph{Deep ensembles (DE).} The \textsc{nam} of \citet{agarwal2021nams} effectively operates as a deep ensemble \citep{lakshminarayanan2017simple} even though the authors do not explicitly present it as such.
In our experiments, we found that ensembles of feature networks using ReLU and GELU activation tend to exhibit a collapse of diversity in function space, as can be seen in the two leftmost panels of \cref{fig:altapproxinf}.
\citet{d2021repulsive} also highlight this as a potential limitation of deep ensembles.
The utilization of ExU activation proposed by \citet{agarwal2021nams} partially restores diversity, and we focus on this configuration when comparing to the \ourmethod.

\paragraph{Mean-field variational inference (MFVI).} Variational inference can also be used to obtain independent approximate feature network posteriors.
In early experiments, we tested the Bayes-by-Backprop (\emph{BbB}) procedure of \citet{blundell2015bbb} and found that the method required significant manual tuning since the feature networks tended to either severely underfit the functions or the uncertainty.
MFVI also yields log-marginal likelihood estimates, however their use in gradient-optimization of neural network hyperparameters appears to be limited.

\paragraph{MC-Dropout.} Dropout \citep{gal2016dropout} can be a simple way of introducing uncertainty awareness in the \textsc{nam} but its Bayesian interpretation is not as straightforward as that of the Laplace approximation.
It requires multiple forward passes for inference and does not provide an inherent mechanism for feature selection.

\subsection{Ablation of the Anion Gap in MIMIC-III}\label{apd:ablation}

The anion gap is a measure of the difference between the serum concentration of sodium and the serum concentrations of chloride and bicarbonate, i.e.~$\mathrm{anion\ gap} = [\mathrm{Na}^{+}] - ([\mathrm{Cl}^{-}] + [\mathrm{HCO_3^{-}]}).$

\begin{figure}[h]
    \centering
    \includegraphics[width=0.8\textwidth]
        {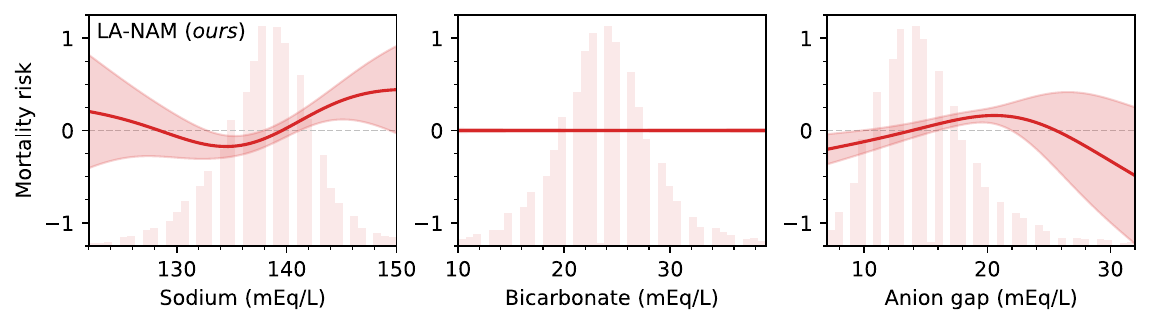}
    \caption[]{Sodium, bicarbonate and anion gap mortality risk as predicted by the \ourmethod.}
    \label{fig:withaniongap}
\end{figure}

Both low bicarbonate levels and thus high anion gap are indicators of acute metabolic acidosis. This is a known risk factor for intensive care mortality with very poor prognosis~\citep{kraut2010acidosis1,kraut2012acidosis2}.
\cref{fig:withaniongap} shows that the predicted mortality risk increases steadily as the anion gap grows but becomes uncertain above 20~mEq/L due to low sample size.

\begin{figure}[h]
    \centering
    \includegraphics[width=0.8\textwidth]
        {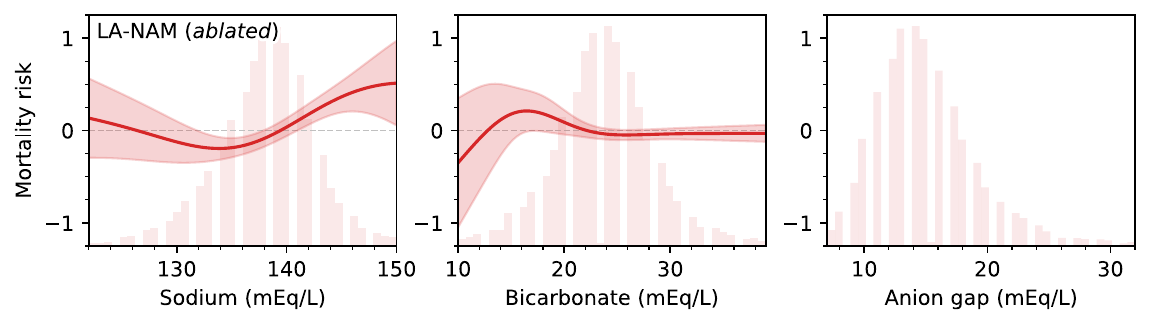}
    \caption[]{Sodium and bicarbonate mortality risk with anion gap feature network ablated.}
    \label{fig:withoutaniongap}
\end{figure}

\noindent When presented with both anion gap and bicarbonate in the mortality risk dataset of \cref{sec:expmimiciii}, the \ourmethod uses high anion gap as a proxy for the risk of low bicarbonate. We confirm this visually by performing an ablation experiment in which the \ourmethod is re-trained with the feature network attending to the anion gap removed.
\cref{fig:withoutaniongap} shows that in the ablated model the anion gap risk is moved into the low levels for bicarbonate. The bicarbonate risk increases below 20~mEq/L and becomes uncertain around 15~mEq/L.

\clearpage
In contrast, the \textsc{oak-gp} does not suppress the bicarbonate, possibly indicating that the orthogonality constraint which it enforces has difficulty addressing redundancy of information when variables are highly correlated.

\begin{figure}[h]
    \centering
    \includegraphics[width=0.8\textwidth]
        {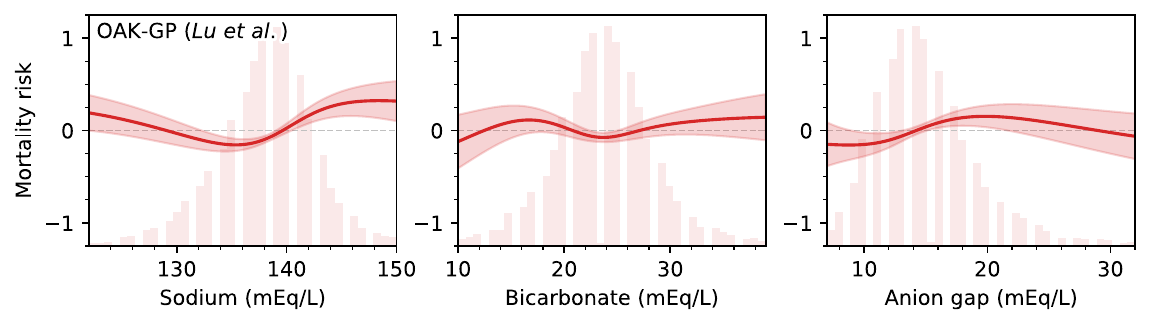}
    \caption[]{Sodium, bicarbonate and anion gap mortality risk in the \textsc{oak-gp} of \citet{lu2022oak}.}
    \label{fig:withaniongapoak}
\end{figure}

\subsection{Ablation of the Activation Function}
\label{apd:depthactiv}

\begin{figure}[!h]
	\includegraphics[width=\textwidth]{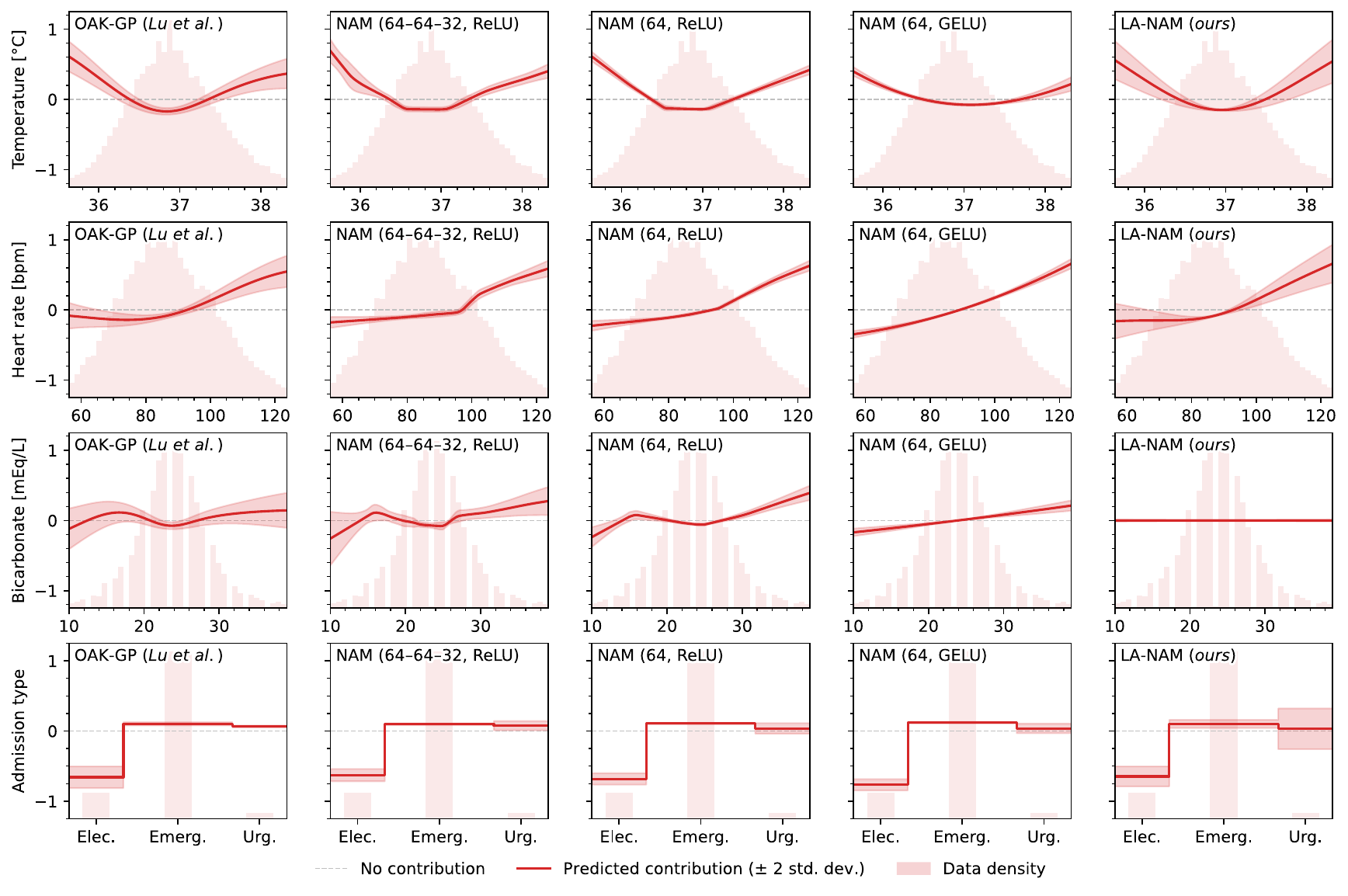}
	\centering
	\caption[]{Progressive ablation of the feature network depth and activation function on the MIMIC-III dataset of \cref{sec:expmimiciii}. Feature network architecture is denoted in parentheses. ``\ourmethod (ours)'' is a single layer of 64 GELU units.}
	\label{fig:depthactiv}
\end{figure}

In \cref{fig:depthactiv} and \cref{tab:depthactiv}, we progressively ablate the feature network depth and activation function  of the \textsc{nam} to match ours.
Shallow networks and GELU activation encourage smoother fits at the expense of worse predictive uncertainty and performance.
This is not a concern for the \ourmethod when applied to the same architecture

\begin{table}[!h]
	\caption{Mean and standard errors over 5 runs for the performance of the ablated models shown in \cref{fig:depthactiv}.}
	\vspace{0.1cm}
	\begin{adjustbox}{max width=0.6\textwidth,center}
	   \begin{tabular}{l|rrrr}
\toprule
MIMIC-III Performance & AUROC ($\uparrow$) & AUPRC ($\uparrow$) & NLL ($\downarrow$) \\
\midrule
\textsc{nam} (64–64–32, ReLU)& 78.89 (±0.04) & 34.30 (±0.04) & 0.2669 (±2e-4) \\
\textsc{nam} (64, ReLU) & 79.02 (±0.02) & 34.28 (±0.02) & 0.2665 (±1e-4) \\
\textsc{nam} (64, GELU) & 77.54 (±0.02) & 34.10 (±0.02) & 0.2699 (±1e-4) \\
\textsc{la-nam} (64, GELU) & \textbf{79.58 (±0.01)} & \textbf{34.77 (±0.04)} & \textbf{0.2644 (±1e-4)} \\
\bottomrule
\end{tabular}

	\end{adjustbox}
	\label{tab:depthactiv}
\end{table}

\subsection{Calibration on the ICU Mortality Prediction Tasks}
\label{apd:medcalib}

\begin{figure}[h]
	\includegraphics[width=0.80\textwidth]{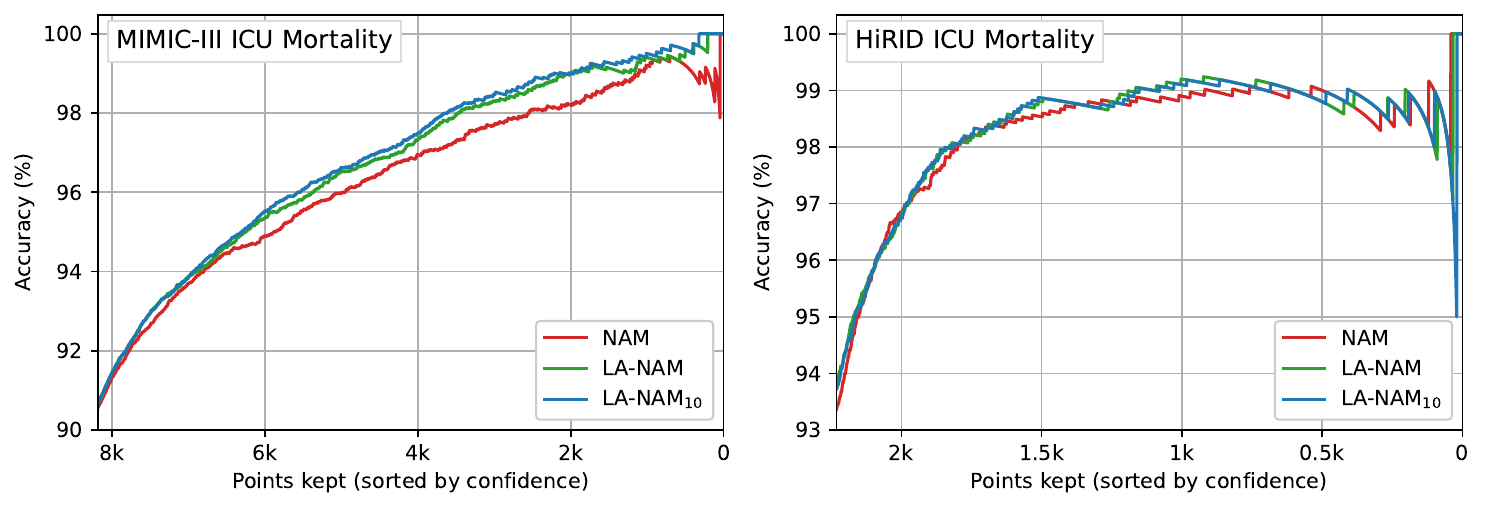}
	\centering
	\caption[]{Calibration curves of the mortality prediction models showing the relationship between uncertainty (horizontal axis) and accuracy (vertical axis). In well-calibrated models, accuracy should increase monotonically.}
	\label{fig:retention}
\end{figure}

We assess the calibration of the \textsc{nam} and \ourmethod predictions on the ICU mortality tasks of \cref{sec:expmimiciii}.
Following the protocol introduced by \citet{ciosek2020conservative}, we present the accuracy as a function of subsets of retained test points in \cref{fig:retention}.
The points are first sorted based on the confidence level of the models and then progressively added to the subset to determine a rolling accuracy.
The median of the rolling accuracy curves for 5 independent runs is shown.
On the MIMIC-III dataset, the \ourmethod demonstrates superior calibration compared to the \textsc{nam}, and it exhibits similar calibration on the significantly larger HiRID ICU dataset. 
Importantly, the incorporation of feature interactions does not seem to adversely harm calibration.

In \cref{tab:medcalib}, we report mean calibration errors and their respective standard errors over the same 5 independent runs.
We provide both the standard expected calibration error (ECE) of \citet{guo2017calibration}, and the root Brier score (RBS), a proper calibration error put forward by \citet{gruber2022better}.
Bold values signify best calibration within one standard error, with the exception of \textsc{la-nam}$_{10}$ where bold indicates on par or better calibration compared to the first-order methods.

\begin{table}[h]
	\caption{Calibration errors over the 5 runs of the mortality models assessed in \cref{fig:retention}.}
	\begin{adjustbox}{max width=\textwidth,center}
		\begin{tabular}{l|rr|rr|rr}
\toprule
 Model & \multicolumn{2}{c}{\textsc{nam}} & \multicolumn{2}{c}{\textsc{la-nam}} & \multicolumn{2}{c}{\textsc{la-nam$_{10}$}} \\
\cmidrule(lr){2-3} \cmidrule(lr){4-5} \cmidrule(lr){6-7}
 Metric & ECE ($\downarrow$) & RBS ($\downarrow$) & ECE ($\downarrow$) & RBS ($\downarrow$) & ECE ($\downarrow$) & RBS ($\downarrow$) \\
\midrule
MIMIC-III ICU Mortality & 0.019 ±3.7e-4 & 0.279 ±1.2e-5 & {\bfseries 0.009 ±1.6e-4} & {\bfseries 0.275 ±2.5e-5} & {\bfseries 0.009 ±6.7e-4} & {\bfseries 0.275 ±9.6e-5} \\
HiRID ICU Mortality & 0.029 ±2.4e-3 & 0.223 ±3.7e-4 & {\bfseries 0.009 ±3.3e-4} & {\bfseries 0.219 ±1.6e-4} & {\color{BrickRed} 0.011 ±4.9e-4} & {\bfseries 0.219 ±2.3e-4} \\
\bottomrule
\end{tabular}

	\end{adjustbox}
	\label{tab:medcalib}
\end{table}

\subsection{Calibration on UCI Datasets}
\label{apd:ucicalib}

We also provide calibration errors and respective standard errors for the 5-fold UCI benchmarks in \cref{tab:ucicalibregr,tab:ucicalibclf}.
For the UCI classification datasets we reuse the metrics of \cref{apd:medcalib}, namely the expected calibration error (ECE) of \citet{guo2017calibration} and the root Brier score (RBS) proposed by \citet{gruber2022better}.
In UCI regression, we provide the quantile error metric proposed by \citet{kuleshov2018accurate} which we denote as ``Calib.''
We also present the squared kernel calibration error for Gaussian predictions (SKCE) proposed by \citet{widmann2022calibration}.
Similarly to \cref{apd:medcalib}, bold denotes best within first-order methods with the exception of \textsc{la-nam}$_{10}$ where bold means on-par or better.

\begin{table}[!h]
    \caption{Calibration error on UCI regression datasets. (Lower is better.)}
    \begin{adjustbox}{max width=\textwidth,center}
        \begin{tabular}{l|rr|rr|rr}
\toprule
 Model & \multicolumn{2}{c}{\textsc{nam}} & \multicolumn{2}{c}{\textsc{la-nam}} & \multicolumn{2}{c}{\textsc{la-nam$_{10}$}} \\
 \cmidrule(lr){2-3} \cmidrule(lr){4-5} \cmidrule(lr){6-7}
 Metric & Calib. ($\downarrow$) & SKCE ($\downarrow$) & Calib. ($\downarrow$) & SKCE ($\downarrow$) & Calib. ($\downarrow$) & SKCE ($\downarrow$) \\
\midrule
autompg ($n = 392$) & {\bfseries 0.054 ±0.011} & {\bfseries 1.9e-3 ±7.0e-4} & {\bfseries 0.037 ±0.007} & {\bfseries 1.8e-3 ±3.5e-4} & {\bfseries 0.043 ±0.009} & {\color{OliveGreen}\bfseries 1.5e-3 ±4.4e-4} \\
concrete ($n = 1030$) & {\bfseries 0.021 ±0.007} & {\bfseries 6.2e-4 ±1.6e-4} & {\bfseries 0.014 ±0.004} & {\bfseries 6.2e-4 ±1.3e-4} & {\bfseries 0.015 ±0.005} & {\bfseries 6.5e-4 ±1.2e-4} \\
energy ($n = 768$) & {\bfseries 0.042 ±0.018} & {\bfseries 8.3e-3 ±4.8e-4} & {\bfseries 0.040 ±0.010} & {\bfseries 8.9e-3 ±4.1e-4} & {\bfseries 0.042 ±0.014} & {\color{OliveGreen}\bfseries 6.1e-3 ±1.3e-3} \\
kin8nm ($n = 8192$) & {\bfseries 0.015 ±0.002} & 5.6e-5 ±1.0e-5 & {\bfseries 0.012 ±0.001} & {\bfseries 2.7e-6 ±4.4e-6} & {\color{OliveGreen}\bfseries 0.007 ±0.001} & {\color{BrickRed} 1.4e-5 ±4.5e-6} \\
naval ($n = 11934$) & 0.176 ±0.006 & 3.1e-8 ±4.4e-9 & {\bfseries 0.077 ±0.029} & {\bfseries 2.7e-9 ±1.5e-10} & {\color{OliveGreen}\bfseries 0.026 ±0.004} & {\color{OliveGreen}\bfseries 2.2e-9 ±1.8e-11} \\
power ($n = 9568$) & {\bfseries 0.005 ±0.002} & 2.3e-4 ±3.1e-5 & {\bfseries 0.005 ±0.002} & {\bfseries 1.9e-4 ±1.3e-5} & {\color{BrickRed} 0.008 ±0.001} & {\color{OliveGreen}\bfseries 1.7e-4 ±1.5e-5} \\
protein ($n = 45730$) & 0.037 ±0.003 & 1.8e-2 ±2.5e-4 & {\bfseries 0.020 ±0.002} & {\bfseries 1.5e-2 ±2.7e-4} & {\color{OliveGreen}\bfseries 0.015 ±0.000} & {\bfseries 1.5e-2 ±1.5e-4} \\
wine ($n = 1599$) & 0.021 ±0.002 & 1.7e-3 ±5.6e-4 & {\bfseries 0.012 ±0.005} & {\bfseries 6.9e-4 ±3.2e-4} & {\bfseries 0.010 ±0.004} & {\bfseries 5.8e-4 ±3.1e-4} \\
yacht ($n = 308$) & {\bfseries 0.095 ±0.027} & {\bfseries 1.8e-2 ±2.0e-3} & {\bfseries 0.105 ±0.028} & {\bfseries 1.7e-2 ±2.3e-3} & {\bfseries 0.111 ±0.024} & {\color{OliveGreen}\bfseries 1.2e-3 ±1.6e-4} \\
\bottomrule
\end{tabular}

    \end{adjustbox}
    \label{tab:ucicalibregr}
\end{table}

\begin{table}[!h]
    \caption{Calibration error on UCI classification datasets. (Lower is better.)}
    \begin{adjustbox}{max width=\textwidth,center}
        \begin{tabular}{l|rr|rr|rr}
\toprule
 Model & \multicolumn{2}{c}{\textsc{nam}} & \multicolumn{2}{c}{\textsc{la-nam}} & \multicolumn{2}{c}{\textsc{la-nam$_{10}$}} \\
\cmidrule(lr){2-3} \cmidrule(lr){4-5} \cmidrule(lr){6-7}
 Metric & ECE ($\downarrow$) & RBS ($\downarrow$) & ECE ($\downarrow$) & RBS ($\downarrow$) & ECE ($\downarrow$) & RBS ($\downarrow$) \\
\midrule
australian ($n = 690$) & 0.112 ±0.008 & {\bfseries 0.330 ±0.017} & {\bfseries 0.099 ±0.004} & {\bfseries 0.314 ±0.014} & {\color{OliveGreen}\bfseries 0.084 ±0.004} & {\bfseries 0.317 ±0.015} \\
breast ($n = 569$) & 0.064 ±0.018 & 0.192 ±0.013 & {\bfseries 0.038 ±0.005} & {\bfseries 0.163 ±0.010} & {\bfseries 0.040 ±0.004} & {\bfseries 0.165 ±0.009} \\
heart ($n = 270$) & {\bfseries 0.153 ±0.008} & 0.356 ±0.020 & {\bfseries 0.136 ±0.011} & {\bfseries 0.314 ±0.014} & {\bfseries 0.150 ±0.015} & {\bfseries 0.315 ±0.015} \\
ionosphere ($n = 351$) & {\bfseries 0.111 ±0.014} & {\bfseries 0.292 ±0.014} & {\bfseries 0.086 ±0.012} & {\bfseries 0.266 ±0.022} & {\bfseries 0.080 ±0.006} & {\bfseries 0.270 ±0.015} \\
parkinsons ($n = 195$) & {\bfseries 0.139 ±0.006} & {\bfseries 0.290 ±0.019} & {\bfseries 0.136 ±0.009} & {\bfseries 0.280 ±0.023} & {\color{OliveGreen}\bfseries 0.111 ±0.006} & {\bfseries 0.277 ±0.025} \\
\bottomrule
\end{tabular}

    \end{adjustbox}
    \label{tab:ucicalibclf}
\end{table}

\subsection{Additional Results on UCI Datasets and ICU Mortality Tasks}
\label{apd:uciresults}

In this section, we report the mean performance and standard errors for additional models and metrics over the 5-fold cross-validation UCI benchmarks of \cref{sec:expucibench} and ICU mortality tasks of \cref{sec:expmimiciii}.
Bolded values indicate best performance within additive models, and {\color{OliveGreen} green} or {\color{BrickRed} red} an improvement or decrease beyond one standard error when second-order interactions are added.
In addition to the models of \cref{tab:uci10nll}, we give performance for linear and logistic regression, the smoothing-spline \textsc{gam}, and the \ourmethod with 10 interactions selected using the improvement to the marginal likelihood lower bound which is discussed in \cref{apd:intermodsel} and denoted as \textsc{la-nam}$_{\mathrm{10}}^\dagger$, instead of the MI-based selection presented in \cref{sec:interaction}.
We also benchmark against the gradient boosting-based \emph{explainable boosting model} (\textsc{ebm}) \citep{lou2012boosting,nori2019interpretml}, which is assigned a maximum likelihood fit of the observation noise in regression using its training data and also supports selection of top-10 feature interaction pairs (denoted as \textsc{ebm}\textsubscript{10}).

\begin{table}[h]
    \caption[]{Performance of the \textsc{ebm} on the mortality tasks of \cref{sec:expmimiciii}.}
    \begin{tabular}{lccccccccc}
\toprule
 Task & \multicolumn{3}{c}{MIMIC-III ICU Mortality} & \multicolumn{3}{c}{HiRID ICU Mortality} \\
 \cmidrule(lr){2-4} \cmidrule(lr){5-7} \cmidrule(lr){8-10}
 Metric & AUROC ($\uparrow$) & AUPRC ($\uparrow$) & NLL ($\downarrow$) & AUROC ($\uparrow$) & AUPRC ($\uparrow$) & NLL ($\downarrow$) \\
\midrule
\textsc{nam} & 77.6 ± 0.03 & 32.3 ± 0.03 & 0.274 ± 8e-5 & 89.6 ± 0.17 & {\bfseries 60.7 ± 0.14} & 0.228 ± 1e-2 \\
\textsc{la-nam} & {\bfseries 79.6 ± 0.01} & 34.8 ± 0.04 & {\bfseries 0.264 ± 5e-5} & 90.1 ± 0.03 & {\bfseries 60.5 ± 0.14} & {\bfseries 0.174 ± 2e-4} \\
\textsc{la-nam}\textsubscript{10} & {\color{OliveGreen} {\bfseries 80.2 ± 0.10}} & {\color{OliveGreen} {\bfseries 35.2 ± 0.06}} & {\color{OliveGreen} {\bfseries 0.262 ± 3e-4}} & 90.1 ± 0.01 & {\bfseries 60.5 ± 0.20} & {\bfseries 0.174 ± 4e-4} \\
\midrule
\textsc{oak-gp} & {\bfseries 79.9 ± 0.03} & {\bfseries 35.2 ± 0.11} & {\bfseries 0.263 ± 1e-4} & --- & --- & --- \\
\textsc{oak-gp}\textsubscript{*} & {\color{BrickRed}71.7 ± 0.66} & {\color{BrickRed}28.5 ± 0.42} & {\color{BrickRed}0.288 ± 1e-3} & --- & ---  & --- \\
\midrule
\textsc{ebm} & 78.7 ± 0.02 & 33.6 ± 0.04 & 0.268 ± 9e-5 & {\bfseries 90.2 ± 0.03} & 59.2 ± 0.10 & 0.177 ± 2e-4 \\
\textsc{ebm}\textsubscript{10} & {\color{OliveGreen} {\bfseries 79.7 ± 0.03}} & {\color{OliveGreen} 34.9 ± 0.09} & {\color{OliveGreen} {\bfseries 0.264 ± 2e-4}} & {\color{OliveGreen} {\bfseries 90.5 ± 0.04}} & {\color{OliveGreen} {\bfseries 61.1 ± 0.23}} & {\color{OliveGreen} {\bfseries 0.173 ± 4e-4}} \\
\midrule
LightGBM & {\bfseries 80.6 ± 0.08} & {\bfseries 35.6 ± 0.19} & {\bfseries 0.261 ± 3e-4} & {\bfseries 90.7 ± 0.00} & {\bfseries 61.6 ± 0.00} & {\bfseries 0.172 ± 0.00} \\
\bottomrule
\end{tabular}

    \centering
\end{table}

\clearpage

\afterpage{
    \clearpage
    \thispagestyle{plain}
    \begin{sidewaystable}[h!]
        \caption{Negative test $\log$-likelihood on UCI regression (top) and classification (bottom). (Lower is better.)}
        \begin{adjustbox}{max width=\textwidth,center}
            \begin{tabular}{l|rrrrrr|rrrr|r}
\toprule
Dataset & Linear & \textsc{gam} & \textsc{nam} & \textsc{la-nam} & \textsc{oak-gp} & \textsc{ebm} & \textsc{la-nam}\textsubscript{10} & \textsc{la-nam}$_{10}^\dagger$ & \textsc{oak-gp}\textsubscript{*} & \textsc{ebm}\textsubscript{10} & LightGBM \\
\midrule
autompg ($n$ = 392) & 2.59 ±0.06 & {\bfseries 2.43 ±0.09} & 2.69 ±0.16 & {\bfseries 2.46 ±0.08} & {\bfseries 2.55 ±0.10} & 2.64 ±0.10 & {\bfseries 2.45 ±0.09} & {\bfseries 2.41 ±0.08} & {\bfseries 2.46 ±0.14} & 2.99 ±0.29 & {\bfseries 2.53 ±0.07} \\
concrete ($n$ = 1030) & 3.78 ±0.04 & {\bfseries 3.13 ±0.05} & 3.46 ±0.12 & 3.25 ±0.03 & {\bfseries 3.19 ±0.09} & {\bfseries 3.20 ±0.12} & {\color{OliveGreen} {\bfseries 3.18 ±0.04}} & {\color{OliveGreen} {\bfseries 3.14 ±0.03}} & {\color{OliveGreen} {\bfseries 2.81 ±0.06}} & {\color{BrickRed} 3.53 ±0.20} & {\bfseries 3.09 ±0.09} \\
energy ($n$ = 768) & 2.46 ±0.02 & {\bfseries 1.46 ±0.02} & {\bfseries 1.48 ±0.02} & {\bfseries 1.44 ±0.02} & {\bfseries 1.46 ±0.02} & {\bfseries 1.46 ±0.02} & {\color{OliveGreen} {\bfseries 1.11 ±0.12}} & {\color{OliveGreen} {\bfseries 1.01 ±0.14}} & {\color{OliveGreen} {\bfseries 0.61 ±0.04}} & {\color{OliveGreen} {\bfseries 0.67 ±0.05}} & {\bfseries 0.81 ±0.05} \\
kin8nm ($n$ = 8192) & -0.18 ±0.01 & {\bfseries -0.20 ±0.01} & -0.18 ±0.01 & {\bfseries -0.20 ±0.00} & 0.09 ±0.01 & {\bfseries -0.20 ±0.01} & {\color{OliveGreen} {\bfseries -0.28 ±0.02}} & {\color{OliveGreen} {\bfseries -0.29 ±0.02}} & 0.09 ±0.01 & {\color{OliveGreen} {\bfseries -0.34 ±0.01}} & {\bfseries -0.50 ±0.03} \\
naval ($n$ = 11934) & -3.72 ±0.01 & -8.09 ±0.02 & -3.87 ±0.01 & -7.24 ±0.01 & {\bfseries -8.93 ±0.07} & -3.15 ±0.00 & {\color{OliveGreen} -7.44 ±0.07} & -7.35 ±0.12 & {\color{OliveGreen} {\bfseries -9.43 ±0.01}} & {\color{OliveGreen} -3.60 ±0.01} & -5.19 ±0.01 \\
power ($n$ = 9568) & 2.94 ±0.01 & 2.84 ±0.01 & 2.89 ±0.02 & 2.85 ±0.01 & {\bfseries 2.81 ±0.03} & {\bfseries 2.79 ±0.02} & {\color{OliveGreen} {\bfseries 2.79 ±0.01}} & {\color{OliveGreen} {\bfseries 2.79 ±0.01}} & {\color{OliveGreen} {\bfseries 2.73 ±0.02}} & {\color{OliveGreen} {\bfseries 2.72 ±0.02}} & {\bfseries 2.67 ±0.02} \\
protein ($n$ = 45730) & 3.06 ±0.00 & 3.02 ±0.00 & 3.02 ±0.00 & 3.02 ±0.00 & {\bfseries 3.00 ±0.00} & {\bfseries 3.00 ±0.00} & {\color{OliveGreen} {\bfseries 2.94 ±0.01}} & {\color{OliveGreen} {\bfseries 2.94 ±0.00}} & {\color{OliveGreen} {\bfseries 2.88 ±0.00}} & {\color{OliveGreen} {\bfseries 2.89 ±0.01}} & {\bfseries 2.83 ±0.00} \\
wine ($n$ = 1599) & {\bfseries 1.00 ±0.03} & {\bfseries 0.98 ±0.03} & {\bfseries 1.02 ±0.04} & {\bfseries 0.98 ±0.03} & 1.14 ±0.04 & {\bfseries 0.99 ±0.03} & {\bfseries 0.97 ±0.03} & {\bfseries 0.97 ±0.03} & 1.67 ±0.65 & {\bfseries 1.01 ±0.04} & {\bfseries 0.96 ±0.03} \\
yacht ($n$ = 308) & 3.64 ±0.07 & {\bfseries 1.87 ±0.10} & 2.24 ±0.08 & {\bfseries 1.81 ±0.10} & {\bfseries 1.86 ±0.11} & {\bfseries 1.93 ±0.13} & {\color{OliveGreen} {\bfseries 0.76 ±0.20}} & {\color{OliveGreen} {\bfseries 1.00 ±0.11}} & {\color{OliveGreen} {\bfseries 0.79 ±0.17}} & {\color{BrickRed} 4.80 ±1.59} & {\bfseries 1.37 ±0.28} \\
\midrule
australian ($n$ = 690) & {\bfseries 0.35 ±0.02} & {\bfseries 0.35 ±0.04} & {\bfseries 0.38 ±0.04} & {\bfseries 0.34 ±0.03} & {\bfseries 0.33 ±0.03} & {\bfseries 0.33 ±0.03} & {\bfseries 0.34 ±0.03} & {\bfseries 0.35 ±0.03} & {\bfseries 0.35 ±0.03} & {\bfseries 0.32 ±0.04} & {\bfseries 0.31 ±0.03} \\
breast ($n$ = 569) & 0.10 ±0.01 & {\bfseries 0.09 ±0.02} & 0.16 ±0.03 & {\bfseries 0.10 ±0.02} & {\bfseries 0.07 ±0.01} & 0.12 ±0.02 & 0.10 ±0.02 & {\bfseries 0.10 ±0.02} & {\bfseries 0.09 ±0.01} & 0.12 ±0.02 & {\bfseries 0.09 ±0.01} \\
heart ($n$ = 270) & {\bfseries 0.39 ±0.04} & 0.43 ±0.08 & 0.41 ±0.04 & {\bfseries 0.33 ±0.02} & 0.42 ±0.06 & 0.40 ±0.02 & {\bfseries 0.33 ±0.03} & {\bfseries 0.34 ±0.03} & 0.42 ±0.06 & 0.41 ±0.03 & {\bfseries 0.39 ±0.04} \\
ionosphere ($n$ = 351) & 0.33 ±0.03 & 0.27 ±0.02 & 0.31 ±0.04 & {\bfseries 0.25 ±0.04} & {\bfseries 0.22 ±0.02} & {\bfseries 0.23 ±0.02} & {\bfseries 0.27 ±0.03} & {\bfseries 0.27 ±0.03} & {\bfseries 0.21 ±0.03} & {\bfseries 0.22 ±0.02} & {\bfseries 0.19 ±0.03} \\
parkinsons ($n$ = 195) & 0.33 ±0.02 & 0.36 ±0.02 & {\bfseries 0.29 ±0.04} & {\bfseries 0.26 ±0.03} & {\bfseries 0.27 ±0.03} & {\bfseries 0.28 ±0.05} & {\bfseries 0.25 ±0.03} & {\bfseries 0.25 ±0.03} & {\bfseries 0.21 ±0.02} & {\bfseries 0.36 ±0.12} & {\bfseries 0.22 ±0.03} \\
\bottomrule
\end{tabular}
        \end{adjustbox}
        \label{tab:uci10nllextra}
        \vspace{1.5cm}
        \caption{Root mean squared error on UCI regression datasets. (Lower is better.)}
        \begin{adjustbox}{max width=\textwidth,center}
            \begin{tabular}{l|rrrrrr|rrrr|r}
\toprule
Dataset & Linear & \textsc{gam} & \textsc{nam} & \textsc{la-nam} & \textsc{oak-gp} & \textsc{ebm} & \textsc{la-nam}\textsubscript{10} & \textsc{la-nam}$_{10}^\dagger$ & \textsc{oak-gp}\textsubscript{*} & \textsc{ebm}\textsubscript{10} & LightGBM \\
\midrule
autompg ($n$ = 392) & 3.18 ±0.18 & {\bfseries 2.70 ±0.20} & {\bfseries 2.94 ±0.19} & {\bfseries 2.77 ±0.18} & {\bfseries 2.84 ±0.15} & {\bfseries 2.98 ±0.11} & {\bfseries 2.72 ±0.19} & {\bfseries 2.66 ±0.17} & {\bfseries 2.70 ±0.23} & {\bfseries 2.86 ±0.15} & {\bfseries 2.96 ±0.13} \\
concrete ($n$ = 1030) & 10.50 ±0.40 & {\bfseries 5.57 ±0.24} & 6.89 ±0.45 & 6.27 ±0.13 & {\bfseries 6.16 ±0.80} & {\bfseries 5.15 ±0.25} & {\color{OliveGreen} 5.80 ±0.19} & {\color{OliveGreen} 5.60 ±0.17} & {\color{OliveGreen} {\bfseries 4.29 ±0.29}} & {\color{OliveGreen} {\bfseries 4.41 ±0.17}} & {\bfseries 4.94 ±0.35} \\
energy ($n$ = 768) & 2.84 ±0.05 & {\bfseries 1.04 ±0.02} & {\bfseries 1.06 ±0.02} & {\bfseries 1.04 ±0.02} & {\bfseries 1.04 ±0.02} & {\bfseries 1.04 ±0.02} & {\color{OliveGreen} {\bfseries 0.76 ±0.08}} & {\color{OliveGreen} {\bfseries 0.68 ±0.09}} & {\color{OliveGreen} {\bfseries 0.44 ±0.02}} & {\color{OliveGreen} {\bfseries 0.47 ±0.02}} & {\bfseries 0.52 ±0.02} \\
kin8nm ($n$ = 8192) & 0.20 ±0.00 & {\bfseries 0.20 ±0.00} & 0.20 ±0.00 & {\bfseries 0.20 ±0.00} & 0.26 ±0.00 & {\bfseries 0.20 ±0.00} & {\color{OliveGreen} {\bfseries 0.18 ±0.00}} & {\color{OliveGreen} {\bfseries 0.18 ±0.00}} & 0.26 ±0.00 & {\color{OliveGreen} {\bfseries 0.17 ±0.00}} & {\bfseries 0.13 ±0.00} \\
naval ($n$ = 11934) & 6e-3 ±4e-5 & 7e-5 ±2e-6 & 5e-3 ±5e-5 & 2e-4 ±2e-6 & {\bfseries 3e-5 ±6e-6} & 1e-2 ±5e-5 & {\color{OliveGreen} 1e-4 ±1e-5} & 2e-4 ±2e-5 & {\color{OliveGreen} {\bfseries 2e-5 ±2e-6}} & {\color{OliveGreen} 7e-3 ±5e-5} & 1e-3 ±6e-6 \\
power ($n$ = 9568) & 4.56 ±0.06 & 4.15 ±0.06 & 4.34 ±0.06 & 4.18 ±0.06 & {\bfseries 4.02 ±0.13} & {\bfseries 3.89 ±0.06} & {\color{OliveGreen} {\bfseries 3.93 ±0.06}} & {\color{OliveGreen} {\bfseries 3.94 ±0.06}} & {\color{OliveGreen} {\bfseries 3.69 ±0.10}} & {\color{OliveGreen} {\bfseries 3.62 ±0.06}} & {\bfseries 3.39 ±0.06} \\
protein ($n$ = 45730) & 5.19 ±0.01 & 4.94 ±0.01 & 4.98 ±0.01 & 4.94 ±0.01 & 4.87 ±0.02 & {\bfseries 4.84 ±0.01} & {\color{OliveGreen} {\bfseries 4.58 ±0.02}} & {\color{OliveGreen} {\bfseries 4.57 ±0.02}} & {\color{OliveGreen} {\bfseries 4.32 ±0.01}} & {\color{OliveGreen} {\bfseries 4.36 ±0.02}} & {\bfseries 4.09 ±0.01} \\
wine ($n$ = 1599) & {\bfseries 0.65 ±0.02} & {\bfseries 0.65 ±0.01} & {\bfseries 0.64 ±0.02} & {\bfseries 0.64 ±0.02} & 0.75 ±0.03 & {\bfseries 0.64 ±0.02} & {\bfseries 0.64 ±0.02} & {\bfseries 0.64 ±0.02} & 0.71 ±0.05 & {\bfseries 0.62 ±0.01} & {\bfseries 0.62 ±0.02} \\
yacht ($n$ = 308) & 9.09 ±0.54 & {\bfseries 1.51 ±0.16} & 2.20 ±0.15 & {\bfseries 1.45 ±0.17} & {\bfseries 1.47 ±0.17} & {\bfseries 1.56 ±0.16} & {\color{OliveGreen} {\bfseries 0.72 ±0.13}} & {\color{OliveGreen} {\bfseries 0.81 ±0.09}} & {\color{OliveGreen} {\bfseries 0.54 ±0.08}} & {\color{OliveGreen} {\bfseries 0.68 ±0.09}} & {\bfseries 0.79 ±0.11} \\
\bottomrule
\end{tabular}
        \end{adjustbox}
        \label{tab:uci10rmseextra}
    \end{sidewaystable}
    \clearpage
}
\afterpage{
    \clearpage
    \thispagestyle{plain}
    \begin{sidewaystable}[h!]
        \caption{Area under the ROC curve in percent on UCI classification datasets. (Higher is better.)}
        \begin{adjustbox}{max width=\textwidth,center}
            \begin{tabular}{l|rrrrrr|rrrr|r}
\toprule
Dataset & Linear & \textsc{gam} & \textsc{nam} & \textsc{la-nam} & \textsc{oak-gp} & \textsc{ebm} & \textsc{la-nam}\textsubscript{10} & \textsc{la-nam}$_{10}^\dagger$ & \textsc{oak-gp}\textsubscript{*} & \textsc{ebm}\textsubscript{10} & LightGBM \\
\midrule
australian ($n$ = 690) & {\bfseries 92.55 ±0.96} & {\bfseries 91.91 ±1.46} & {\bfseries 92.01 ±1.04} & {\bfseries 92.60 ±1.14} & {\bfseries 93.43 ±0.99} & {\bfseries 93.17 ±1.27} & {\bfseries 92.55 ±1.11} & {\bfseries 92.14 ±1.01} & {\bfseries 93.41 ±1.01} & {\bfseries 93.39 ±1.35} & {\bfseries 93.80 ±1.00} \\
breast ($n$ = 569) & {\bfseries 99.60 ±0.24} & {\bfseries 99.40 ±0.35} & 98.97 ±0.38 & {\bfseries 99.45 ±0.21} & {\bfseries 99.61 ±0.24} & {\bfseries 99.24 ±0.19} & {\bfseries 99.42 ±0.21} & {\bfseries 99.39 ±0.28} & {\bfseries 99.56 ±0.26} & {\bfseries 99.35 ±0.21} & {\bfseries 99.30 ±0.30} \\
heart ($n$ = 270) & {\bfseries 90.00 ±2.37} & 89.14 ±2.79 & {\bfseries 90.31 ±2.03} & {\bfseries 93.53 ±1.42} & 89.22 ±2.27 & 90.31 ±1.44 & {\bfseries 93.36 ±1.46} & {\bfseries 93.36 ±1.53} & 89.25 ±2.35 & 89.00 ±1.56 & {\bfseries 91.00 ±2.00} \\
ionosphere ($n$ = 351) & 90.42 ±2.10 & {\bfseries 95.38 ±0.88} & {\bfseries 95.05 ±1.25} & {\bfseries 94.46 ±1.33} & {\bfseries 96.07 ±0.72} & {\bfseries 96.33 ±0.84} & {\bfseries 94.66 ±1.13} & {\bfseries 94.34 ±1.15} & {\bfseries 96.22 ±0.94} & {\bfseries 97.34 ±0.68} & {\bfseries 97.30 ±0.90} \\
parkinsons ($n$ = 195) & 90.01 ±2.23 & 88.54 ±2.64 & {\bfseries 94.62 ±1.38} & {\bfseries 94.52 ±1.74} & {\bfseries 94.17 ±1.27} & {\bfseries 94.64 ±1.95} & {\bfseries 94.01 ±1.87} & {\bfseries 94.52 ±1.74} & {\color{OliveGreen} {\bfseries 96.43 ±0.80}} & {\color{OliveGreen} {\bfseries 97.41 ±0.73}} & {\bfseries 95.50 ±1.30} \\
\bottomrule
\end{tabular}
        \end{adjustbox}
        \label{tab:uci10aurocextra}
        \vspace{1.5cm}
        \caption{Area under the precision-recall curve in percent on UCI classification datasets. (Higher is better.)}
        \begin{adjustbox}{max width=\textwidth,center}
            \begin{tabular}{l|rrrrrr|rrrr|r}
\toprule
Dataset & Linear & \textsc{gam} & \textsc{nam} & \textsc{la-nam} & \textsc{oak-gp} & \textsc{ebm} & \textsc{la-nam}\textsubscript{10} & \textsc{la-nam}$_{10}^\dagger$ & \textsc{oak-gp}\textsubscript{*} & \textsc{ebm}\textsubscript{10} & LightGBM \\
\midrule
australian ($n$ = 690) & {\bfseries 91.17 ±0.76} & {\bfseries 91.09 ±1.17} & {\bfseries 90.01 ±1.21} & {\bfseries 92.08 ±1.22} & {\bfseries 91.42 ±1.19} & {\bfseries 91.82 ±1.41} & {\bfseries 91.93 ±1.33} & {\bfseries 91.68 ±1.20} & {\bfseries 91.29 ±1.28} & {\bfseries 92.12 ±1.39} & {\bfseries 92.80 ±1.30} \\
breast ($n$ = 569) & {\bfseries 99.72 ±0.18} & {\bfseries 99.59 ±0.26} & 99.19 ±0.39 & {\bfseries 99.63 ±0.15} & {\bfseries 99.75 ±0.16} & {\bfseries 99.52 ±0.13} & {\bfseries 99.62 ±0.16} & {\bfseries 99.59 ±0.19} & {\bfseries 99.72 ±0.18} & {\bfseries 99.59 ±0.14} & {\bfseries 99.50 ±0.20} \\
heart ($n$ = 270) & {\bfseries 88.52 ±2.98} & {\bfseries 87.77 ±3.72} & {\bfseries 89.08 ±2.83} & {\bfseries 92.58 ±1.78} & {\bfseries 88.46 ±3.40} & 88.71 ±1.88 & {\bfseries 92.56 ±1.85} & {\bfseries 92.51 ±1.92} & {\bfseries 88.41 ±3.41} & 87.85 ±1.94 & {\bfseries 89.10 ±3.20} \\
ionosphere ($n$ = 351) & 91.05 ±2.19 & {\bfseries 97.16 ±0.60} & {\bfseries 96.28 ±1.08} & 95.44 ±1.19 & {\bfseries 97.22 ±0.61} & {\bfseries 97.44 ±0.72} & 95.11 ±1.32 & 94.87 ±1.36 & {\bfseries 97.42 ±0.72} & {\bfseries 98.21 ±0.52} & {\bfseries 98.10 ±0.70} \\
parkinsons ($n$ = 195) & 96.73 ±0.76 & 95.90 ±1.25 & {\bfseries 98.14 ±0.50} & {\bfseries 98.15 ±0.60} & {\bfseries 98.16 ±0.40} & {\bfseries 98.09 ±0.83} & {\bfseries 98.04 ±0.59} & {\bfseries 98.18 ±0.59} & {\color{OliveGreen} {\bfseries 98.88 ±0.25}} & {\color{OliveGreen} {\bfseries 99.20 ±0.23}} & {\bfseries 98.40 ±0.50} \\
\bottomrule
\end{tabular}

        \end{adjustbox}
        \label{tab:uci10auprcextra}
    \end{sidewaystable}
    \clearpage
}

\clearpage
\subsection{Experimental Setup}\label{apd:expdetails}

\paragraph{Linear/Logistic Regression.} Implementations are provided by the \texttt{scikit-learn} library \citep{pedregosa2011sklearn}. The regularization strength is grid-searched in the set $\{0.001, 0.01, 0.1, 1, 10\}$. For logistic regression, we use the L-BFGS solver performing up to 10,000 iterations.

\paragraph{GAM.} We use an open source Python implementation \citep[\texttt{pygam};][]{serven2018pygam}. The smoothing parameters are grid-searched: We sample one thousand candidates uniformly from the recommended range of $[10^{-3},\,10^3]$ and select using generalized cross-validation scoring \citep[GCV;][]{golub1979gcv}. 

\paragraph{NAM.} We test the NAM with feature networks containing a single hidden layer of 1024 units and ExU activation. We grid-search the learning rate and regularization hyperparameters using the values 
given in the supplementary material of \citet{agarwal2021nams}. In practice, we find that combining dropout with a probability of 0.2 and feature dropout of 0.05 along with weight decay of $10^{-5}$ and a learning rate of either 0.01 or 0.001 gave good results.

\paragraph{LA-NAM.} The \ourmethod is constructed using feature networks containing a single hidden layer of 64 neurons with GELU activation \citep{hendrycks2016gelu}.
Joint feature networks added for second-order interaction fine-tuning contain two hidden layers of 64 neurons.
The feature network parameters and hyperparameters (prior precision, observation noise) are optimized using Adam~\citep{kingma2014adam}, alternating between optimizing both at regular intervals, as in \citet{immer2021scalable}.
We select the learning rate in the discrete set of $\{0.1,\,0.01,\,0.001\}$ which maximizes the ultimate $\log$-marginal likelihood.
We use a batch size of 512 and perform early stopping on the $\log$-marginal likelihood restoring the best scoring parameters and hyperparameters at the end of training.
We find that the algorithm is fairly robust to the choice of hyperparameter optimization schedule: For all experiments, we use $0.1$ for the hyperparameter learning rate and perform batches of 30 gradient steps on the $\log$-marginal likelihood every 100 epochs of regular training.

\paragraph{OAK-GP.} We use the setup recommended by the authors \citep[See Appendix H. of][]{lu2022oak}. Sparse GP regression is used for large UCI regression datasets ($n \geq 1000$). For the UCI classification and ICU mortality prediction tasks variational inference is used, with 200 inducing points for UCI and 800 inducing points for the mortality prediction.

\paragraph{EBM.} The explainable boosting machine (\textsc{ebm}) is an open-source Python implementation of the gradient-boosting GAM that is available as a part of the \texttt{InterpretML} library \citep{nori2019interpretml}. The library defaults for the hyperparameters performed best. We did not find a significant improvement when tuning the learning rate, maximum number of leaves, or minimum number of samples per leaf.

\paragraph{LightGBM.} We use the open-source implementation \citep{ke2017lightgbm}. The maximum depth of each tree is selected in the set $\{3,\,7,\,12\}$, and maximum number of leaves in the set $\{8,\,16,\,31\}$. Additionally, the minimum number of samples per leaf is reduced to 2 on small datasets from the default of 20. Early stopping is enabled in all experiments, using a 12.5\% split of the training data when the task has no predefined validation data. For the HiRID ICU Mortality task in \cref{tab:medical}, we obtain best performance when using the feature engineering recommended in \citet{yeche2021hiridicubenchmark}.

\paragraph{Hardware.}
The deep learning models are trained on a single \texttt{NVIDIA RTX2080Ti} with a \texttt{Xeon E5-2630v4} core.
Other models are trained on \texttt{Xeon E5-2697v4} cores and \texttt{Xeon Gold 6140} cores.

\clearpage

\subsection{Illustrative Example}\label{apd:toyexample}
\renewcommand{\textfraction}{0.0} %

\begin{figure}[!h]
    \centering
    \includegraphics[width=0.99\textwidth]
        {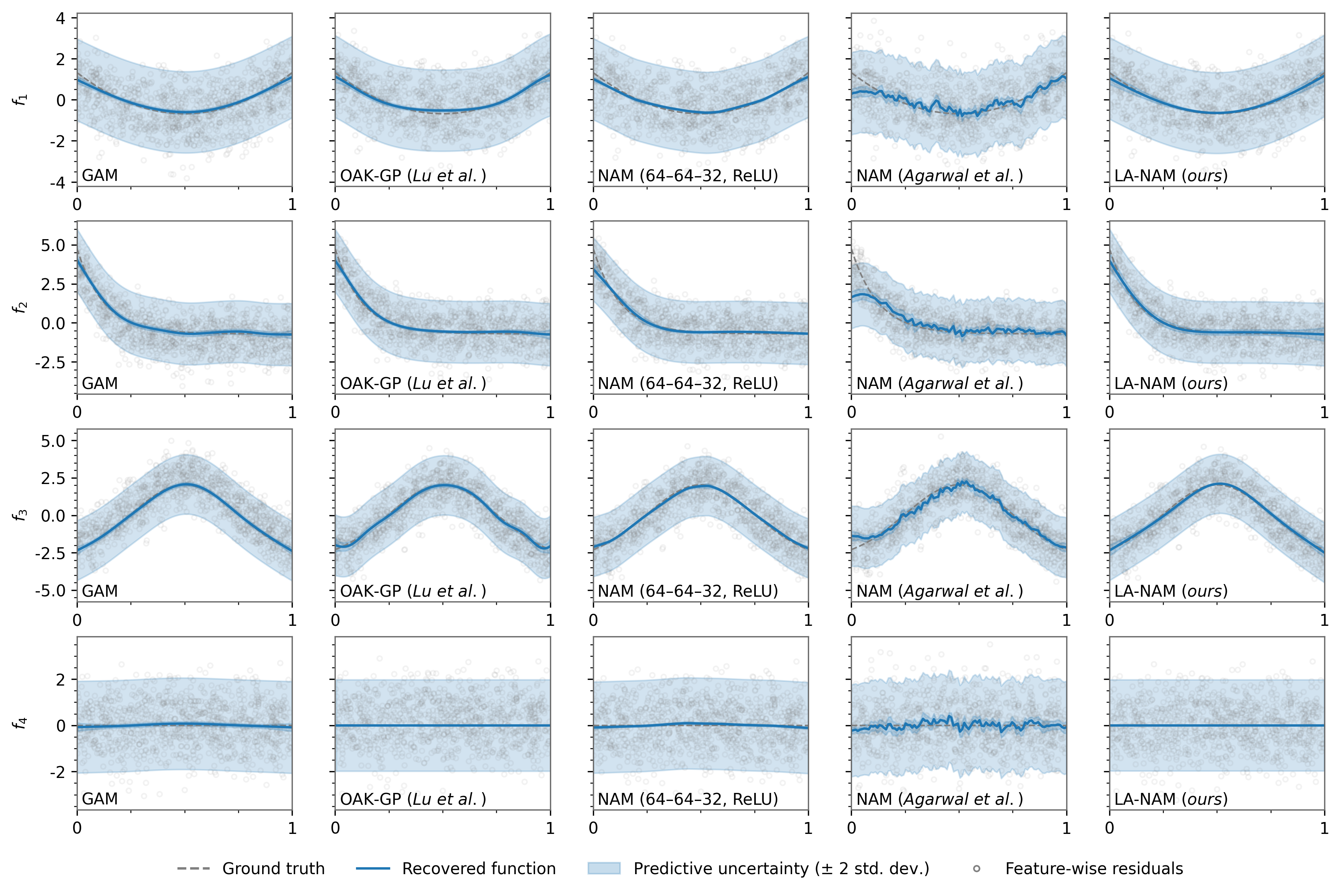}
    \vspace{-0.4cm}
    \caption[]{Recovery of the additive structure of the synthetic dataset of \ref{apd:toyexample}. The feature-wise residuals are the generated data points with mean contribution of the other feature networks subtracted.}
    \label{fig:toyexamplelarge}
\end{figure}

In \cref{sec:exptoyexample}, we present an illustrative example to motivate the capacity of the \ourmethod and baselines to recover purely additive structure from noisy data.
We provide further details on the generation of the synthetic dataset used here.
Consider the function $\hat{f} : \mathbb{R}^4 \to \mathbb{R}$, where $\hat{f}(x_1,\,x_2,\,x_3,\,x_4) = \hat{f}_1(x_1) + \hat{f}_2(x_2) + \hat{f}_3(x_3) + \hat{f}_4(x_4)$, and
\begin{align}
    &   \hat{f}_1(x_1) = 8 (x_1 - \tfrac{1}{2})^2, \quad
    &&  \hat{f}_2(x_2) = \tfrac{1}{10} \exp[-8 x_2 + 4] \\
    &   \hat{f}_3(x_3) = 5 \exp[-2(2 x_3 - 1)^2], \quad
    &&  \hat{f}_4(x_4) = 0.
\end{align}
We generate $N = 1000$ noisy observations $\{\mathbf{x}_n,\,y_n\}_{n=1}^N$ by sampling inputs $\mathbf{x}_n$ uniformly from $\mathcal{U}([0,\,1]^4)$ and generating targets $y_n = \hat{f}(\mathbf{x}_n) + \epsilon_n$, where $\epsilon_n \sim \mathcal{N}(0,\,1)$ is random Gaussian noise. \cref{fig:toyexamplelarge} shows the recovered functions along with the associated predictive uncertainty for the \ourmethod and baseline models.

\renewcommand{\textfraction}{0.2} %

\clearpage

\subsection{Predicted Mortality Risk in MIMIC-III}\label{apd:mimiciii}
\renewcommand{\textfraction}{0.0} %

\begin{figure}[h]
    \centering
    \includegraphics[
        width=0.99\textwidth %
    ]{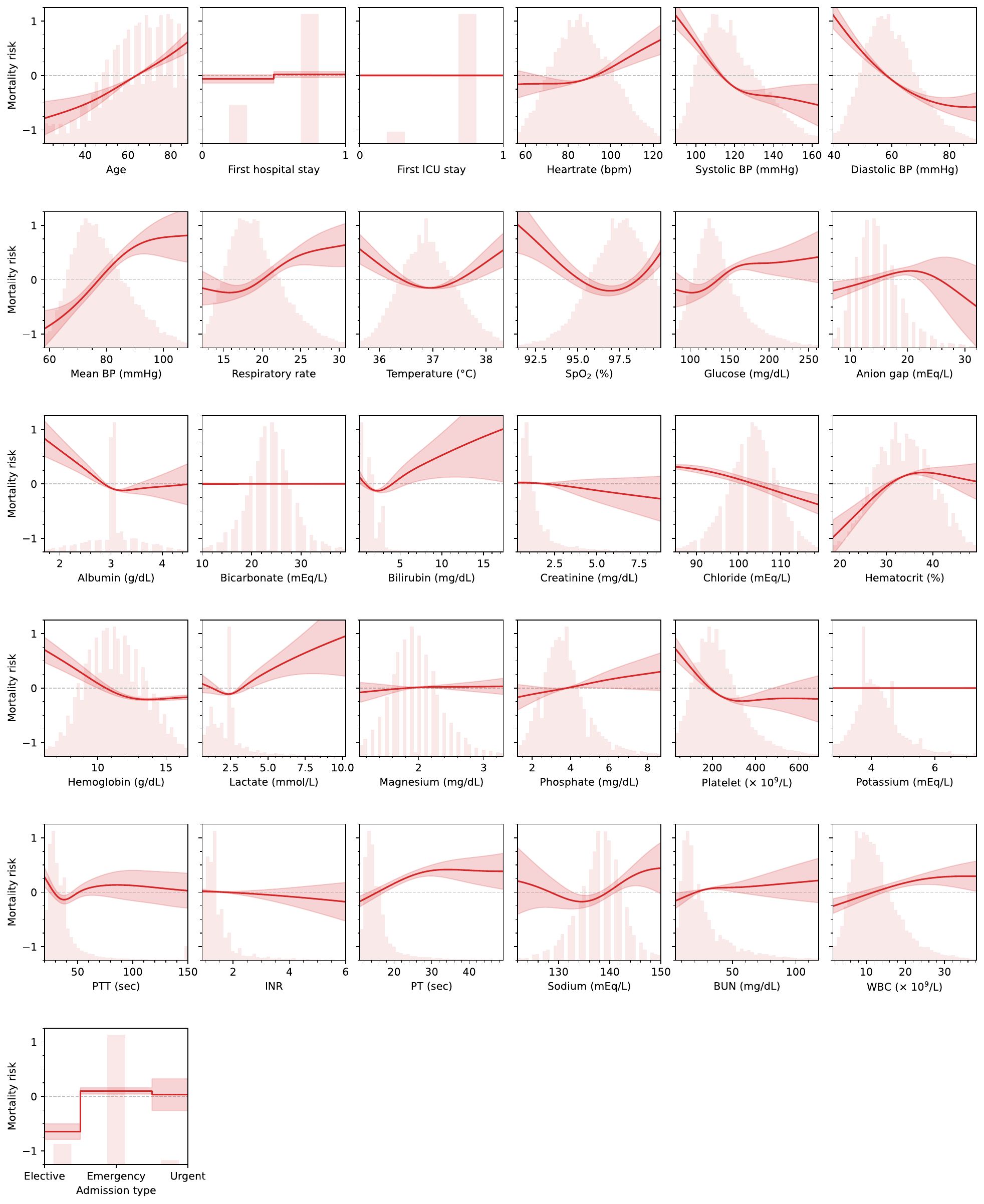}
    \caption[]{Complete visualization of the \ourmethod predicted risk in the MIMIC-III mortality task of \cref{sec:expmimiciii}.}
\end{figure}
\renewcommand{\textfraction}{0.2} %

\newpage

\section{Detailed Related Work}
\label{sec:detail_related_work}

\paragraph{Generalized additive models.}

Generalized additive models have been extensively studied and various approaches have been proposed for their construction.
\citet{hastie1999gams} initially suggested using smoothing splines \citep{wahba1983smoothing} to build the smoothing functions which attend to each feature, fitting them iteratively through ``backfitting'' \citep{breiman1985backfitting}.
One alternative is to construct them using gradient-boosted decision trees \citep{friedman2001pdp}.
By modifying the boosting algorithm, it becomes possible to cycle through functions in the inner loop, which has been found to be favorable compared to sequential backfitting of boosted trees \citep{lou2012boosting}.
Furthermore, boosted trees facilitate the selection and fitting of feature interactions \citep{lou2013gatwom}.
These second-order interactions are believed to contribute to the competitive accuracy achieved by gradient-boosted additive models in comparison to fully interacting models for tabular supervised learning \citep{caruana2015gatwom,nori2019interpretml}.

\paragraph{Sparse additive models.}

Sparse adaptations of the generalized additive models have also been proposed.
In the sparse additive models (SpAM) of \citet{liu2007spam}, component functions are assigned coefficients which are subjected to an $L_1$ constraint and optimized with coordinate descent.
Alternatively, \citet{zhao2012sam} suggested sparse additive machines (SAM) which replace the linear discriminant equation of $L_1$ support vector machines ($L_1$-SVM) with an additive structure of univariate functions.
Group sparsity can be obtained within both formulations \citep{yin2012groupspam,chen2017groupsam}.
More recently, \citet{wang2020mam} have proposed multi-task additive models (MAM) as robust extensions of GroupSpAM with mode regression, which do not require \emph{a priori} knowledge of the sparse feature groups.

\paragraph{Neural additive models.}

Neural networks are highly attractive for constructing smoothing functions due to their ability to approximate continuous functions with arbitrary precision given sufficient complexity  \citep{cybenko1989theorem, maiorov1999theorem, lu2017theorem}.
\citet{agarwal2021nams} proposed the neural additive model (NAM), which utilizes ensembles of feed-forward networks and employs standard backpropagation for fitting.
To accommodate jagged functions, they introduced ``ExU'' dense layers, wherein weights are learned in logarithmic space.

A closely related model, called GAMI-Net, was introduced by \citet{yang2021gaminet}, but it employs single networks instead of an ensemble and additionally supports the learning of feature interaction terms.
In the GAMI-Net, the feature pair candidates are selected for feature interaction fine-tuning using the ranking procedure of \citet{lou2013gatwom}. \citet{yang2021gaminet} acknowledge that this ranking can also be done using neural networks with the approach proposed by \citet{tsang2018nid}.
Alternatively, one can avoid having to select pairs entirely by finding a scalable formulation for all possible pairs, such as in the neural basis model (NBM) of \citet{radenovic2022nbm}.
In this work, we propose to rank the feature pair candidates using the available posterior approximation.

Many other extensions have been suggested, including feature selection through sparse regularization of the feature networks \citep{xu2022sparsenam}, generation of confidence intervals using regression spline basis expansion \citep{luber2023structnam}, and estimation of the skewness, heteroscedasticity, and kurtosis of the underlying data distributions \citep{thielmann2023namlss}.

\paragraph{Bayesian neural networks.}
Bayesian neural networks offer the potential to combine the expressive capabilities of neural networks with the rigorous statistical properties of Bayesian inference \citep{mackay1992practical,neal1993bayesian}.
They have recently sparked increased attention \citep{arbel2023primer} and it has been claimed that their capabilities will be crucial in generative AI \citep{manduchi2024challenges} and modern large-scale AI in general \citep{papamarkou2024position}.
However, achieving accurate inference in these complex models has proven to be a challenging endeavor \citep{jospin2022hands}.
The field has explored various techniques for approximate inference, each with its own trade-offs in terms of quality and computational cost.

At one end of the spectrum, we have inexpensive local approximations such as Laplace inference \citep{laplace1774approx, mackay1992practical, khan2019approximate, daxberger2021redux}, which provides a simple and computationally efficient solution.
Other approaches in this category include stochastic weight averaging \citep{izmailov2018averaging,maddox2019simple} and dropout \citep{gal2016dropout,kingma2015variational}.

Moving towards more sophisticated approximations, variational methods come into play, offering a diverse range of complexity levels and approximations \citep{graves2011vi,blundell2015bbb,louizos2016structured,khan2018vogn,osawa2019practical}.
Moreover, ensemble-based methods have also been explored as an alternative avenue \citep{lakshminarayanan2017simple, wang2019function, ciosek2020conservative, d2021stein, d2021repulsive}.

On the other end of the spectrum, we find Markov Chain Monte Carlo (MCMC) approaches, which provide asymptotically correct solutions.
Many recent works have contributed to the exploration of these computationally expensive yet theoretically attractive methods \citep{neal1993bayesian, welling2011bayesian, garriga2021exact, izmailov2021hmc}.

Beyond the challenges related to approximate inference, recent work has also studied the question of prior choice for Bayesian neural networks \citep[e.g.,][and references therein]{fortuin2021bnnpriors, fortuin2022bayesian, fortuin2022priors, nabarro2022data, sharma2023incorporating}.
Additionally, model selection within the Bayesian neural network framework has garnered attention \citep[e.g.,][]{immer2021scalable, immer2022invariance, immer2022inductive, rothfuss2021pacoh, rothfuss2022pac, van2022learning, schwobel2022last}, which we use to select features and regularize the networks automatically.

\paragraph{Modern Laplace Approximations.} 
In our work, we focus primarily on leveraging the linearized Laplace inference technique and the associated marginal likelihood estimate~\citep{mackay1992practical}.
While sophisticated methods can enable posterior (predictive) inference, the Laplace approximation is one of the few scalable methods that offer a useful marginal likelihood approximation for neural networks~\citep{immer2021scalable}.
To the best of our knowledge, it has not been previously applied to the NAM.
Automatic relevance determination, which we use to select features, has been previously used in similar contexts for fully-connected networks~\citep{mackay1994bayesian}, regularization of individual outputs~\citep{immer2023effective}, or more general architecture selection~\citep{van2023learning}.

\end{document}